\documentclass[runningheads]{llncs}

\usepackage{eccv}

\usepackage{eccvabbrv}      
\usepackage{graphicx}
\usepackage{booktabs}
\usepackage[accsupp]{axessibility}  

\usepackage{hyperref}

\newcommand{\secref}[1]{\S\ref{#1}}
\newcommand{\figref}[1]{Figure~\ref{#1}}
\newcommand{\tabref}[1]{Table~\ref{#1}}

\newcommand{\sysname}{\textsc{CooperScene}\xspace}
\newcommand{\Sysname}{\textsc{CooperScene}\xspace}

\definecolor{goodgreen}{RGB}{204,255,204}
\definecolor{mediumyellow}{RGB}{255,255,204}
\definecolor{badred}{RGB}{255,204,204}

\usepackage[utf8]{inputenc} 
\usepackage[T1]{fontenc}    
\usepackage{url}            
\usepackage{booktabs}       
\usepackage{amsfonts}       
\usepackage{nicefrac}       
\usepackage{microtype}      
\usepackage{xcolor}         
\usepackage{wrapfig}
\usepackage{xspace}
\usepackage{subcaption}

\usepackage{booktabs,multirow,xcolor,colortbl}
\usepackage{multirow}
\usepackage{xcolor}
\usepackage{colortbl}
\usepackage[normalem]{ulem}
\usepackage{balance}
\usepackage{siunitx}

\newcommand\paraspace{\vspace*{0.5ex}}
\providecommand\parab[1]{\paraspace\noindent\textbf{#1}}

\begin{document}

\title{\Sysname: Multi-Modal Cooperative Autonomy Benchmark
with C-V2X Communication Characterization\vspace{-5mm}}
\titlerunning{\Sysname}

\author{
Bo Wu$^{\ast}$ \and
Ruoshen Mo$^{\ast}$ \and
Justin Yue \and
Yanyu Zhang \and
Janice Nguyen \and
Guoyuan Wu \and
Amit Roy-Chowdhury \and
Matthew J. Barth \and
Hang Qiu$^\dagger$}
\authorrunning{B. Wu et al.}
\institute{University of California, Riverside, USA \\
\scriptsize\{bwu109, rmo008, jyue009, yzhan831, jnguy172, guoyuanw, amitrc, barth, hangq\}@ucr.edu}
\maketitle
\renewcommand{\thefootnote}{}
\footnotetext{$\ast$ Equal contribution. $\dagger$ Corresponding author.}
\renewcommand{\thefootnote}{\arabic{footnote}}

\vspace{-5mm}
\begin{center}
    \centering
    \includegraphics[width=0.83\textwidth]{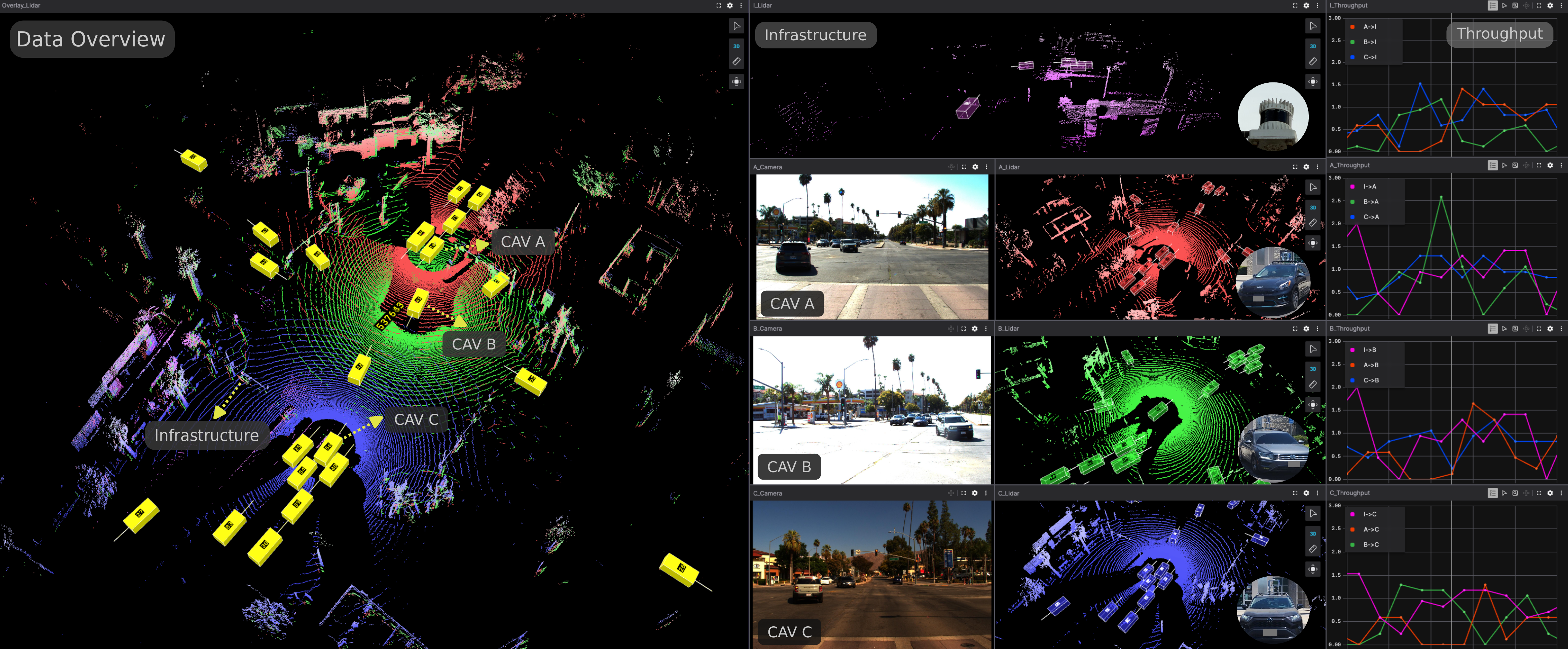}
    \begin{minipage}{0.83\textwidth}
    \captionof{figure}{\Sysname sample data visualization. 
    Left: Overlay of point clouds from all agents (three connected  autonomous vehicles (CAVs) in red, green, blue, and infrastructure in purple) with global 3D object labels (yellow). 
    Middle: Individual agent sensor views in local frames.
    Right: Real time C-V2X throughput measured between all agent pairs.}
    \label{fig:teaser}
    \end{minipage}
\end{center}
\vspace{-11mm}

\begin{abstract}
Cellular vehicle-to-everything (C-V2X) enables cooperative perception, prediction, and planning beyond the field of view of individual agents.
However, existing datasets often overlook the complexities of real-world deployment, such as limited communication bandwidth and its dynamics, heterogeneous sensing modalities, and scalability beyond a single cooperative partner.
In this paper, we introduce \Sysname, a high-fidelity cooperative autonomy dataset with \textit{real-world C-V2X communication characterization}. The dataset is organized into diverse \textit{scenes}, including intersections, highway ramps, and parking lots. These scenes involve three connected and autonomous vehicles (CAVs) and one infrastructure roadside unit (RSU), all equipped with multi-modal sensors and commercial off-the-shelf C-V2X communication radios.
All scenes are annotated with \textit{globally consistent} 3D labels at 10 Hz, totaling \textit{344K objects across 59K frames},
underpinned by \textit{tight sensor- and agent-synchronization}, \textit{centimeter-level localization and spatial alignment}, \textit{precise cross-modality calibration}, and \textit{3GPP-standard-compliant C-V2X communication}. 
\Sysname establishes a rigorous benchmark for evaluating multi-agent scaling and actual performance in real-world deployable settings. 
Project
website for data and benchmark: \url{https://cisl.ucr.edu/CooperScene}.

\end{abstract}



\section{Introduction}
Today's autonomous vehicles (AVs)~\cite{waymo, robotaxi} rely on a combination of onboard multimodal sensors to operate safely and efficiently. However, these autonomous agents are deployed as solitary entities that often suffer from incomplete situation awareness, making unilateral decisions, and stranding in long-tail events, waiting for remote operation~\cite{waymo_remote}.
As AVs become increasingly deployed, these limitations underscore the urgent need for a paradigm shift from \textit{individual to cooperative autonomy}~\cite{sae_j3216,  dot2018preparing, national2020business}.

Cellular-vehicle-to-everything (C-V2X) technology breaks the solitary barrier by enabling vehicles to communicate with other vehicles (V2V), infrastructure (V2I) and vulnerable road users (V2P). V2X connectivity is estimated to prevent up to 595,000 vehicle crashes annually in the U.S., resulting in an annual economic savings of approximately \$71 billion~\cite{national2016department}. 
Recent V2X-enabled cooperative perception~\cite{harbor, v2vnet, xu2022cobevt}, prediction~\cite{wang2025cmp}, and driving~\cite{Cui_2022, CoopReflect, univ2x} techniques emerge as a promising solution to address individual autonomy limitations. Driven by public cooperative perception datasets and benchmarks~\cite{xu2022opv2v, xu2023v2v4real,Sun_2020_CVPR, kitti, nuscenes2019}, these approaches have proven capable of overcoming occlusions, extending sensor range, and improving robustness in dynamic and unstructured environments~\cite{v2xvit,avr}.


\begin{wrapfigure}{r}{0.43\textwidth} 
    \vspace{-6mm} 
    \centering
    \includegraphics[width=\linewidth]{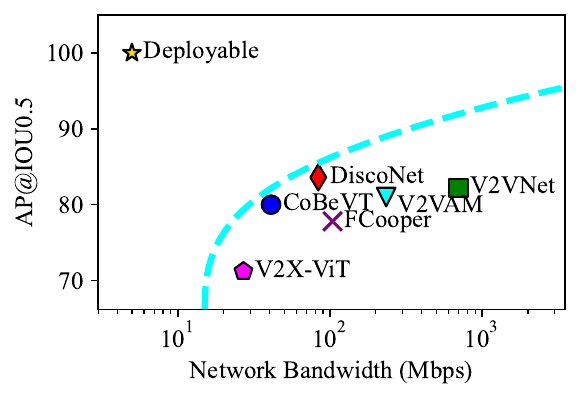}
    \vspace{-6mm}
    \caption{Accuracy and Bandwidth of Open-source Cooperative Perception Methods Benchmarked on OPV2V~\cite{xu2022opv2v} Dataset: A Research Gap towards Real-world Deployment.}
    \label{fig:v2v_bw}
    \vspace{-6mm} 
\end{wrapfigure}

Despite the success on benchmarks, many models fall short in real-world deployment, partly due to a lack of datasets that reflect realistic V2X communication dynamics. \figref{fig:v2v_bw} summarizes the network bandwidth requirements of a few publicly benchmarked models. To achieve reasonable accuracy, the amount of latent features exchanged is often \textit{orders of magnitude} higher than what is available today in C-V2X technology~\cite{seev2x}\footnote{Measured by off-the-shelf C-V2X radios compliant with 3GPP Release 14~\cite{3gpp_v14}} for individual agents in the shared Intelligent Transportation Systems (ITS) spectrum.

Moreover, most existing methods assume homogeneous sensing modalities, which limits their applicability to only a subset of traffic participants: those agents equipped with similar sensors. Meanwhile, a majority of publicly available \textit{real-world} datasets offer only \textit{one} cooperative agent (vehicle or infrastructure) besides the ego agent (see \tabref{tab:coop_datasets}, column \textit{Agent}), which also limits the benchmarking of multi-agent cooperation at scale.

\begin{table*}[t]
\centering
\caption{Overview of \textit{real-world} cooperative perception datasets in several key attributes, including sensing configuration (agents and modalities), localization and synchronization mechanisms, calibration availability, dataset scale, and supported benchmark tasks. 
\textcolor{green}{Green} cells indicate full support or strong capability, \textcolor{yellow}{yellow} denotes partial support, and \textcolor{red}{red} marks missing features. \sysname uniquely integrates C-V2X communication characterization into diverse multi-modal, multi-agent scenarios, establishing a rigorous benchmark for evaluating \textbf{\textit{deployable performance}} in real-world settings.}

\vspace{-3mm}
\label{tab:coop_datasets}

\resizebox{\textwidth}{!}{

\begin{tabular}{ll|cc|ccc|ccc|c|cc|cc|cccc|cc}
\toprule
\multirow{2}{*}{Dataset} &
\multirow{2}{*}{} &
\multicolumn{2}{c}{Agent} &
\multicolumn{3}{c}{Modality} &
\multicolumn{3}{c}{Scenario} &
\multicolumn{1}{c}{Localization} &
\multicolumn{2}{c}{Synchronization} &
\multicolumn{2}{c}{Calibration} &
\multicolumn{4}{c}{Statistics} &
\multicolumn{2}{c}{Task} \\

\cmidrule(lr){3-4}
\cmidrule(lr){5-7}
\cmidrule(lr){8-10}
\cmidrule(lr){11-11}
\cmidrule(lr){12-13}
\cmidrule(lr){14-15}
\cmidrule(lr){16-19}
\cmidrule(lr){20-21}

& & Veh. & Infra. 
& Cam. & LiDAR & V2X 
& Junc. & Ramp & Lot
& Method 
& Sensor & Agent 
& Intr. & Extr. 
& Pcd & Img & Seq & Obj 
& Det. & Mot. \\
\midrule

DAIR-V2X-C~\cite{dair-v2x} & & \cellcolor{badred}{1} & \cellcolor{goodgreen}{1} &
\cellcolor{goodgreen}{\checkmark} & \cellcolor{goodgreen}{\checkmark} &  \cellcolor{badred}{-} &
\cellcolor{goodgreen}{\checkmark} & \cellcolor{badred}{-} & \cellcolor{badred}{-} &
\cellcolor{badred}{GPS+SLAM} & \cellcolor{badred}{-} & \cellcolor{badred}{-} &
\cellcolor{mediumyellow}{\checkmark} & \cellcolor{mediumyellow}{\checkmark} &
\cellcolor{mediumyellow}{39K} & \cellcolor{mediumyellow}{39K} & \cellcolor{goodgreen}{100} & \cellcolor{goodgreen}{464K} &
\cellcolor{goodgreen}{\checkmark} & \cellcolor{badred}{-} \\

V2V4REAL~\cite{xu2023v2v4real} & & \cellcolor{mediumyellow}{2} & \cellcolor{badred}{0} &
\cellcolor{goodgreen}{\checkmark} &\cellcolor{goodgreen}{\checkmark} & \cellcolor{badred}{-} & 
\cellcolor{goodgreen}{\checkmark} & \cellcolor{badred}{-} & \cellcolor{badred}{-} &
\cellcolor{badred}{GPS/IMU} & \cellcolor{badred}{-} & \cellcolor{badred}{-} &
\cellcolor{goodgreen}{\checkmark} & \cellcolor{goodgreen}{\checkmark} &
\cellcolor{mediumyellow}{20K} & \cellcolor{mediumyellow}{40K} & \cellcolor{mediumyellow}{41} & \cellcolor{goodgreen}{240K} &
\cellcolor{goodgreen}{\checkmark} & \cellcolor{badred}{-} \\

V2X-Seq~\cite{v2x-seq} &  & \cellcolor{badred}{1} & \cellcolor{goodgreen}{1} &
\cellcolor{goodgreen}{\checkmark} & \cellcolor{goodgreen}{\checkmark} & \cellcolor{badred}{-} & 
\cellcolor{goodgreen}{\checkmark} & \cellcolor{badred}{-} & \cellcolor{badred}{-} &
\cellcolor{badred}{GPS+SLAM} & \cellcolor{badred}{-} & \cellcolor{badred}{-} &
\cellcolor{mediumyellow}{\checkmark} & \cellcolor{mediumyellow}{\checkmark} &
\cellcolor{mediumyellow}{15K} & \cellcolor{mediumyellow}{15K} & \cellcolor{goodgreen}{95} & \cellcolor{badred}{10K} &
\cellcolor{goodgreen}{\checkmark} & \cellcolor{goodgreen}{\checkmark} \\

TUM-Traf~\cite{tumtraf} &  & \cellcolor{badred}{1} & \cellcolor{goodgreen}{1} &
\cellcolor{goodgreen}{\checkmark} & \cellcolor{goodgreen}{\checkmark} & \cellcolor{badred}{-} & 
\cellcolor{goodgreen}{\checkmark} & \cellcolor{badred}{-} & \cellcolor{badred}{-} &
\cellcolor{mediumyellow}{RTK+ICP} & \cellcolor{mediumyellow}{ROS} & \cellcolor{mediumyellow}{NTP} &
\cellcolor{goodgreen}{\checkmark} & \cellcolor{goodgreen}{\checkmark} &
\cellcolor{badred}{2K} & \cellcolor{badred}{5K} & \cellcolor{mediumyellow}{30} & \cellcolor{badred}{30K} &
\cellcolor{goodgreen}{\checkmark} & \cellcolor{badred}{-} \\

\midrule
\textbf{\sysname} &  &
\cellcolor{goodgreen}{3} & \cellcolor{goodgreen}{1} &
\cellcolor{goodgreen}{\checkmark} & \cellcolor{goodgreen}{\checkmark} & \cellcolor{goodgreen}{\checkmark} & \cellcolor{goodgreen}{\checkmark} & 
\cellcolor{goodgreen}{\checkmark} & \cellcolor{goodgreen}{\checkmark} &
\cellcolor{goodgreen}{RTK+ST-ICP} & \cellcolor{goodgreen}{PTP} & \cellcolor{goodgreen}{GNSS} &
\cellcolor{goodgreen}{\checkmark} & \cellcolor{goodgreen}{\checkmark} &
\cellcolor{goodgreen}{59K} & \cellcolor{goodgreen}{53K} & \cellcolor{mediumyellow}{24} & \cellcolor{goodgreen}{344K} &
\cellcolor{goodgreen}{\checkmark} & \cellcolor{goodgreen}{\checkmark} \\

\bottomrule
\end{tabular}


}
\vspace{-4mm}
\end{table*}

To address these limitations, in this paper, we introduce \sysname, a multi-modal cooperative autonomy benchmark with C-V2X communication characterization. 
The dataset is organized into diverse \textit{``scenes''}, each involving \textit{three connected and autonomous vehicles} and \textit{one infrastructure roadside unit}, all equipped with \textit{multi-modal sensors} and commercial off-the-shelf \textit{C-V2X communication radios}, interacting with each other across diverse real-world traffic scenarios, including urban intersections, freeway, local street, and parking areas, which introduce different traffic and occlusion patterns.
To provide a high-fidelity benchmark, \Sysname is built with \textit{tight sensor- and agent-synchronization}, \textit{centimeter-level localization and spatial alignment}, \textit{precise cross-modality calibration}~\cite{bevcalib}, and \textit{3GPP-standard-compliant C-V2X communication}. 
All scenes are annotated at a frequency of 10 Hz with 3D bounding boxes in a globally consistent frame of reference, totaling approximately \textit{344K labels across 59K frames}, along with synchronized pair-wise C-V2X throughput measurements.
\Sysname benchmarks both the performance and bandwidth of state-of-the-art cooperative perception methods to raise awareness of the research gap in both perception accuracy and communication efficiency. 
%
The summary of the contribution is as~follows:

\vspace{-3mm}

\begin{itemize}
    \item \textbf{Unique integration of C-V2X in realistic cooperative scenes.}
    We introduce high-quality multi-agent interactive scenes collected from three vehicles and one infrastructure unit across diverse real-world scenarios. Each scene contains synchronized C-V2X measurements (latency, throughput, packet loss rate, and jitter) with multi-modal sensor streams (LiDAR, camera, and GNSS-RTK) and globally consistent 3D object labels.
   


    \item \textbf{Data quality refinement and validation.}
    We build \Sysname with careful system design to guarantee high-quality data in terms of \textit{localization, synchronization, and calibration}, using spatial-temporal ICP (ST-ICP), Precision Time Protocol (PTP), and motion capture systems and eye-hand calibration. We introduce a data validation procedure verifying the efficacy of these methods and the quality of the data collected.

    \item \textbf{Deployable cooperative autonomy benchmark.} 
    Using synchronous C-V2X measurements, we provide the actual \textit{deployable performance} benchmark for both detection and motion prediction tasks. In contrast to huge performance gains with infeasible communication assumptions, our practical benchmark raises awareness of the significant challenges in balancing model efficacy and communication efficiency.
\end{itemize}
\vspace{-3mm}

\Sysname serves as a cross-disciplinary benchmark to bridge the vision, robotics, and networking communities, paving the way towards realistic field-deployable cooperative autonomy.

\section{Related Work}
\noindent\textbf{Cooperative Perception Datasets.}
\tabref{tab:coop_datasets} summarizes  \textit{real-world} cooperative perception datasets in contrast with \Sysname.
V2V4REAL~\cite{xu2023v2v4real} provides real-world cooperative perception benchmarks featuring multi-modal data from two vehicles. While it demonstrates the clear benefits of vehicle-to-vehicle (V2V) collaboration, it lacks infrastructure-side sensing. 
In contrast, DAIR-V2X-C~\cite{dair-v2x} and V2X-Seq~\cite{v2x-seq} incorporate a roadside unit (RSU) but pair it with only a single vehicle, thereby limiting cooperation to dual-agent settings. Furthermore, projecting the 3D point cloud into the 2D camera frame in these datasets reveals distinct spatial misalignments, where corresponding objects appear visibly shifted between the two modalities. Crucially, all three datasets report a large inter-agent synchronization error of 30–50~ms without providing a rigorous validation procedure.
%
TUM-Traf~\cite{zimmer2024tumtraf} introduces a relatively smaller vehicle–infrastructure dataset but with high-precision localization; it incorporates an RSU with multi-modal sensors and refined poses via GNSS RTK and the Iterative Closest Point (ICP) algorithm~\cite{ICP}—a geometric registration method that iteratively minimizes the distance between two point clouds to determine their optimal alignment. Nevertheless, it remains limited to dual-agent settings. 

While these datasets have significantly advanced cooperative perception research, several open challenges remain. Most benchmarks involve only one or two participating vehicles, limiting the study of multi-agent cooperation. Additionally, the precision of calibration, synchronization, and localization in these datasets is insufficient, potentially causing temporal drift and inter-agent and cross-modality misalignment.
Recent efforts have begun to address multi-agent scaling: UrbanIng-V2X~\cite{urbaningv2x2025} collects data from multiple vehicles and infrastructure units across several intersections, substantially increasing the number of cooperative agents, though its multi-modal sensing remains limited. 
%
Most importantly, none of the existing datasets record real-world C-V2X throughput, leaving open questions about actual \textit{achievable cooperative gains} due to latency, bandwidth, and packet loss in practical deployment settings.

\Sysname addresses these gaps by providing three vehicles and one infrastructure in the same scene, enabling richer multi-agent interactions with realistic C-V2X communication characterization. The dataset includes multi-modal sensing (LiDAR, camera, GNSS/IMU), precise time synchronization via PTP and hardware triggering, and fully calibrated intrinsic and extrinsic parameters for all sensors. \Sysname is, to our knowledge, the first real-world dataset to provide synchronized C-V2X communication traces alongside multi-agent multi-modal sensing, allowing joint analysis of perception quality and real communication performance. The dataset supports cooperative 3D detection and motion prediction, allowing comprehensive evaluation and benchmark.

\parab{Cooperative Perception Methods.}
Existing cooperative perception methods can be broadly categorized by the type of feature fusion and the underlying network architecture. Early fusion methods~\cite{autocast, emp, avr, rao, cooper}  directly share raw sensor data, such as point clouds or images, among connected agents, but they have substantial communication overhead and require strict synchronization. 
Late fusion~\cite{latefusion1}, on the other hand, aggregates the final object-level detections, which reduces communication cost but sacrifices spatial reasoning.
Intermediate fusion~\cite{v2vnet, xu2022cobevt, harbor, unisense, disconet} approaches strikes a middle ground by sharing compressed or encoded feature maps, enabling efficient collaboration with less bandwidth requirements. 
For example, 
V2VNet~\cite{v2vnet} represents vehicles as nodes and dynamically learns spatial message passing through edges, enabling agents to reason about occluded regions using neighboring information.
V2X-ViT~\cite{v2xvit} introduces a hybrid transformer that alternates between intra-agent self-attention and inter-agent cross-attention to capture both local and global dependencies across agents.
CoBEVT~\cite{xu2022cobevt} further leverages Swin Transformers for multi-scale BEV fusion, achieving fine-grained spatial alignment.
V2VAM~\cite{v2vam} integrates intra-agent and inter-agent attention to dynamically weigh feature importance under lossy communication.
These transformer-based frameworks improve robustness and scalability, at the cost of higher compute and data sharing volume.

The volume of these intermediate representations, while smaller than the raw data, remains well beyond what vehicular networks can support in real time. \figref{fig:v2v_bw} summarizes the performance and required bandwidth from these approaches~\cite{v2vnet, xu2022cobevt, v2vam, disconet, v2xvit, fcooper}. To address this bottleneck, recent communication-efficient methods~\cite{CooperTrim, hu2022where2comm} explicitly optimize the bandwidth demand through feature selection: ERMVP~\cite{ermvp} compresses shared features via filter-and-merge sampling, while CoSDH~\cite{cosdh} performs supply-demand-aware region selection within a hybrid fusion scheme. However, what exacerbates the problem is the stringent decision deadline in these safety-critical autonomous systems. Extra latency incurred by data compression~\cite{xu2022cobevt, draco}, selection~\cite{CooperTrim, hu2022where2comm}, or codebook retrieval~\cite{codebook} further ages the stale data and risks missing the decision deadline.

Beyond perception, methods such as CMP~\cite{wang2025cmp}, UniV2X~\cite{univ2x}, Coopernaut~\cite{Cui_2022}, and CoopReflect~\cite{CoopReflect} extend cooperation to prediction and end-to-end driving.
They jointly learn perception and planning using shared multi-agent observations, demonstrating the potential of cooperative gains in downstream autonomy tasks.

\section{\Sysname Dataset}
In this section, we present \Sysname, a multi-agent multi-modal dataset with C-V2X communication characterization. \Sysname is collected across diverse real-world traffic scenarios. Specifically, the dataset contains 16 intersection scenes, 4 highway ramp scenes, and 4 parking lot scenes.  The dataset features four participating agents: three vehicles equipped with \textbf{C}ooperative \textbf{M}ulti-Modal \textbf{S}ensing platform (CMS) and one instrumented intersection infrastructure system. \Sysname comprises 59K synchronized LiDAR frames and 53K image frames, annotated with 344K 3D bounding boxes, alongside GNSS, IMU, and C-V2X network traces. We provide details in sensing and communication platform setup, synchronization and calibration procedures, spatial alignment methodology, and data collection and annotation processes in this section.

\subsection{Platform Setup}

\noindent\textbf{Cooperative Multi-Modal Sensing (CMS).}  To collect multi-modal
sensor data, we design and build a CMS platform, a networked open-source portable universal sensing hub~\cite{cms}. CMS provides an
extensible platform with streamlined integration of multi-modal sensors, abstracting away the intricacies of calibration and synchronization, and it is easily reproducible and replicable to jump-start multi-purpose robot and multi-agent research.

\begin{wrapfigure}{r}{0.6\columnwidth}
\centering
\vspace{-6mm}
\includegraphics[width=0.6\columnwidth]{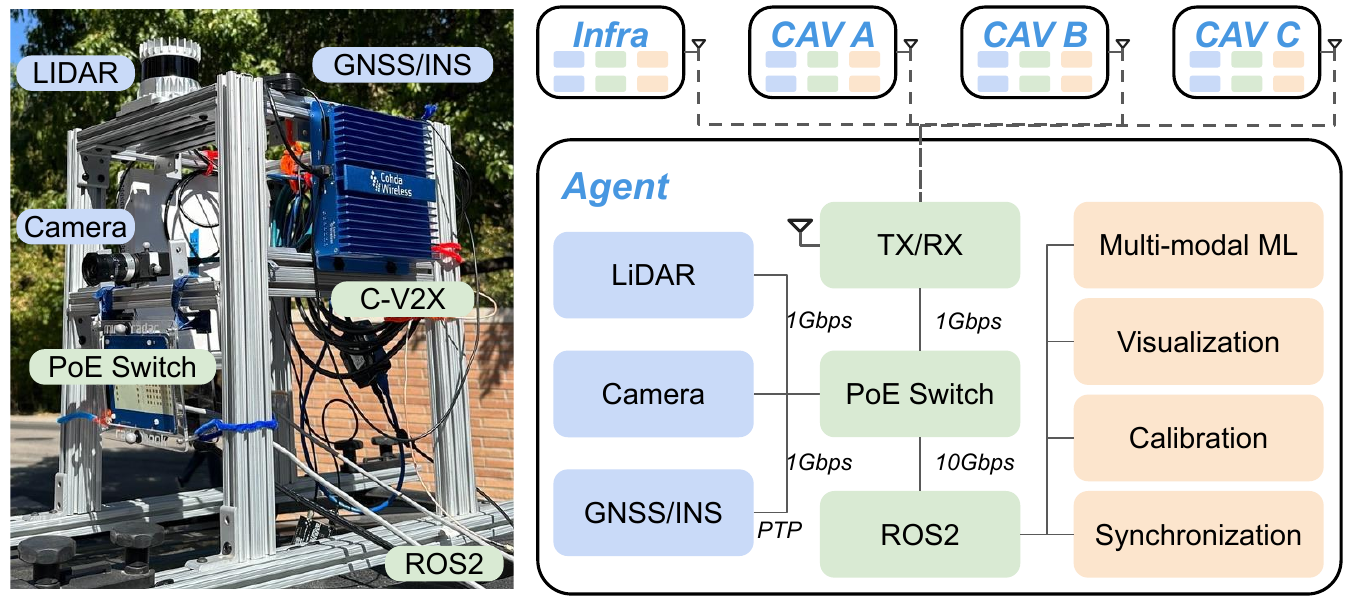}
\caption{CMS Overview. CMS integrates LiDAR, camera, GNSS with a power-over-ethernet (PoE) switch, which forwards the data to a central ROS node (running on a laptop). The laptop and sensors are synchronized with GNSS time, and all intrinsic and extrinsic parameters are calibrated for all sensors. TX/RX module communicates with other CMS platforms on both vehicles and the infrastructure.}
\label{fig:arch}
\centering
\vspace{-6mm}
\end{wrapfigure}
\figref{fig:arch} shows CMS architecture. Specifically, the various sensors are connected to a central ROS node (\eg, running on a PC or laptop) via a power-over-ethernet (PoE) switch. The PoE switch not only redirects the sensor stream via the ROS bus, but also provides power to various sensors. Due to the massive amount of data live-streamed, a switch with Small Form-factor Pluggable (SFP+) port is used to support data transfer rates of up to 10 Gbps. The GNSS module provides microsecond-level accuracy pulse-per-second (PPS) signals to synchronize all sensors (\secref{sec:sensor_sync}). The transmitting/receiving (TX/RX) module enables CMS platforms to communicate with other platforms using vehicle-to-everything (V2X) communication technology.
For \sysname data collection, each CMS platform is equipped with an Ouster OS1-128 LiDAR~\cite{Ouster}, a Lucid Triton 5.4 MP GigE camera~\cite{Lucid}, a TX/RX module (Cohda MK6~\cite{Cohda}) with cellular, WiFi, C-V2X transceivers and GNSS receivers, and a high-resolution Xsens MTi-680 RTK GNSS/INS~\cite{xsense} module. The CMS platform can be mounted on the vehicle~roof.

\parab{Infrastructure Instrumentation.} To capture a complementary infrastructure perspective, we deploy a similar CMS setup (except camera) at a busy intersection. 
One LiDAR sensor is mounted on a traffic light pole, and a GNSS-disciplined MK6 unit synchronizes the LiDAR clocks and communicates with vehicles.
The consistent setup allows us to use the same data collection and processing procedures as those used by the CMS platform on the vehicle.

\subsection{Sensor and Agent Synchronization}
\label{sec:sensor_sync}
Synchronization is critical for data integrity and consistency, as asynchronous data mismatches in time domain and confuses downstream processes. However, depending on the available synchronization support from the sensors, the task can become non-trivial. We first tackle \textit{onboard} synchronization and then discuss \textit{cross-platform} synchronization.

\parab{Onboard Sensor Synchronization.} Although ROS provides built-in time synchronization by timestamping sensor data when the central node receives published messages, it is inaccurate for multi-modal sensors due to network overhead and operating system scheduling. Moreover, each sensor has its own local oscillator offset, leading to asynchronous data collection. 
To address this issue, we use PTP~\cite{PTP} to synchronize sensor clocks and apply hardware triggering to ensure LiDAR and camera data are captured simultaneously. In CMS platform, Cohda MK6 acts as the PTP master and disciplines all other devices, including the LiDAR, camera, and the laptop.
Beyond synchronizing the local clocks, to further guarantee synchronized capture between the LiDAR and camera, we use a General-Purpose Input/Output (GPIO) pin from the LiDAR to generate a stable 10 Hz synchronization pulse that triggers the camera to capture images. Specifically, the camera captures an image upon detecting a rising edge of the pulse, which is generated at the same time the LiDAR begins data collection.

\parab{Cross-platform Agent Synchronization.} With onboard sensors synchronized, the next challenge is to align time across multiple platforms. Achieving cross-platform synchronization requires a common global time reference. Prior work has explored both centralized signals (\eg, WiFi~\cite{WiFiSync}, cellular~\cite{cellSync}) and distributed protocols~\cite{RBS,FTSP} in various contexts.
For outdoor autonomous driving scenarios, where GNSS signals are available, we use the GNSS signal to discipline the Cohda MK6. Cohda MK6 then serves as a grandmaster clock to synchronize the local clocks of onboard sensors with PTP support. Because each agent's grandmaster clock is locked to the same global GNSS time, all sensors across all platforms achieve accurate global time alignment.

\subsection{Spatial Alignment}
\label{sec:icp}


\begin{wrapfigure}{r}{0.45\columnwidth}
\vspace{-8mm}
\includegraphics[width=0.45\columnwidth]{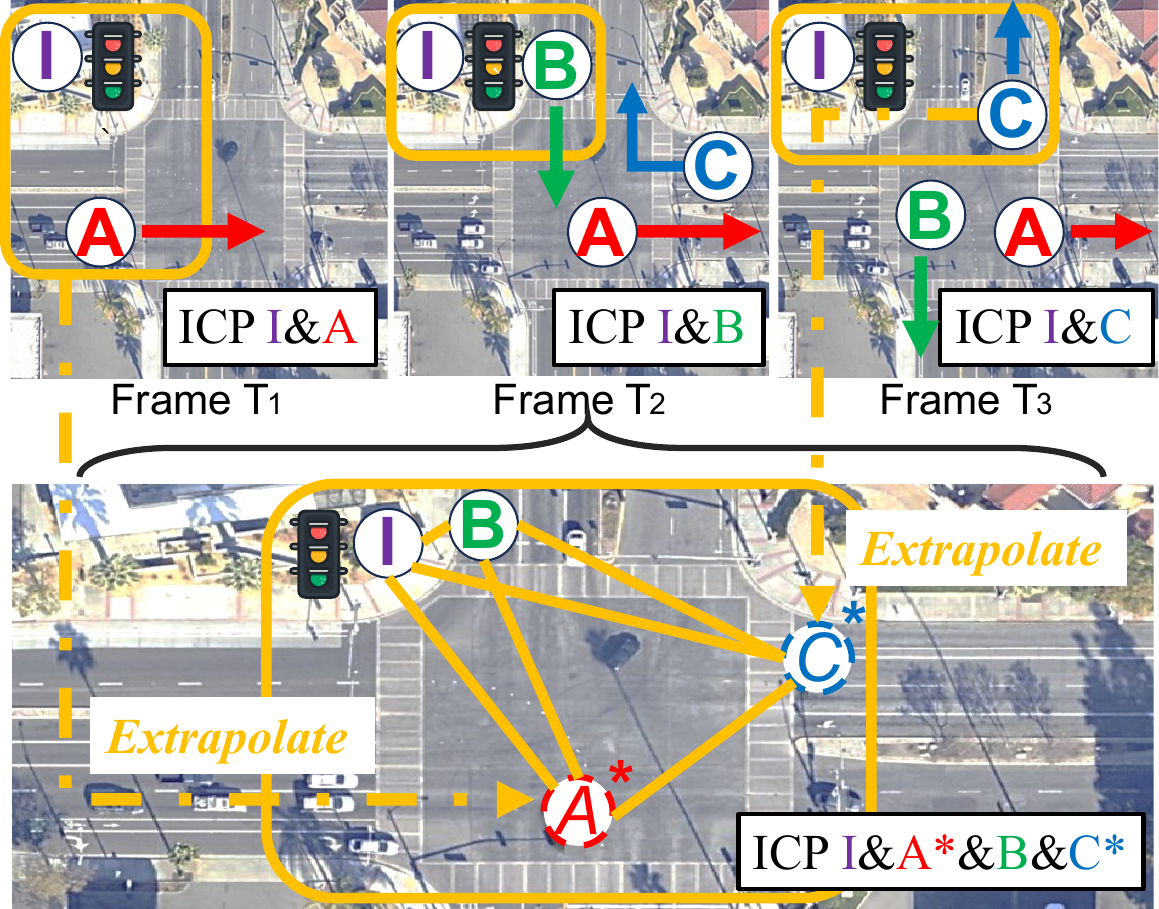}
\caption{Multi-agent Spatial-temporal Pose Alignment}
\label{fig:icp_figure}
\vspace{-6mm}
\end{wrapfigure}

GNSS/IMU errors and noises in real-world systems often lead to spatial misalignment among LiDAR frames collected from different agents.
ICP~\cite{ICP} is widely used to align two point clouds to mitigate this problem.
However, standard ICP cannot be directly applied to the multi-agent scenarios in \sysname. To address this, we propose a spatial-temporal ICP approach, as illustrated in Fig.~\ref{fig:icp_figure}.
First, we use GNSS positions and IMU orientations to project all LiDAR frames from each agent’s local coordinate system into a global East-North-Up (ENU) frame centered at the intersection. This step computes the transformation matrix $T_{local}^{global}$.
Next, for each vehicle, we search its entire sequence to find the frame with the maximum LiDAR overlap with the infrastructure's LiDAR. We run ICP on this single, optimal frame pair to find a coarse adjustment transform $T_{PairwiseICP}$.
Finally, given the high accuracy and temporal stability of GNSS-RTK positioning, the relative distance changes within the same agent remain consistent across frames. We therefore extrapolate the adjustment found in each vehicle-infra pair to a frame where all agents are present and closest to each other; the frame that minimizes the total inter-agent distance. In this extrapolated ``joint'' frame, we perform another round of ICP to simultaneously align all vehicles and the infrastructure, obtaining the global transformation $T_{JointICP}$.
Each vehicle’s final spatial correction matrix is computed as:
$T = T_{JointICP}\times T_{PairwiseICP}\times T_{local}^{global}$.
After spatial-temporal ICP alignment, all vehicles are aligned with the infrastructure and mutually aligned with one another.

\subsection{Sensor Calibration}
To better leverage the complementary nature of different sensing modalities, one key insight in multi-modal sensing is to use the returns of different sensors from the same object to enhance perception accuracy and confidence. The task of sensor calibration is to figure out the intrinsic and extrinsic parameters of all sensors, such that the downstream processing pipeline can understand the \textit{correspondence} of the data from each modality. The device intrinsic (\eg camera focal length and principal point) is unique to each sensor. The calibration methods to obtain these intrinsics are well-studied~\cite{intrinsics_tsai,intrisics_zhang}. In this section, we focus on extrinsic (relative pose) calibration~\cite{bevcalib} between multi-modal sensors. Existing methods typically fall into two categories: target-based approaches~\cite{zhang2004} that rely on predefined geometries (\eg, checkerboards or AprilTags), and targetless, data-driven approaches~\cite{bevcalib, calibnet}. However, these conventional techniques either focus on a single modality pair or fail to achieve sufficient alignment precision.

To get groundtruth-level pair-wise calibration, one approach is to use a motion capture system (MoCap)~\cite{optitrack, vicon}, putting multiple markers on each device to get the pose estimate. While MoCap gives millimeter-level accuracy, the limitation of this approach is that the marker placement outside the sensor cannot represent the optical model inside the sensor, hence making it difficult to calculate the origin of each sensor's coordinate system. The issue can be rectified by solving eye-hand calibration~\cite{4059150}. For a particular sensor, we formulate the eye-hand problem as solving for $X$ in $AX=XB$, where $A$ and $B$ are the relative motions between the markers' poses in MoCap ($T_{markers}^{mocap}$) and the transforms from a fiducial marker to the sensor. Eye-hand then yields $T_{sensor}^{markers}$, the transform from a sensor's optical center to the markers, and we perform the following calculation to find the true pose of the sensor in MoCap: $T_{sensor}^{mocap} = T_{markers}^{mocap} * T_{sensor}^{markers}$.

After finding the true sensor poses, we can accurately calculate the extrinsics between sensors. Given sensor poses $T_{S_1}$ and $T_{S_2}$ for sensors $S_1$ and $S_2$, the extrinsic from $S_1$ to $S_2$ is $T_{S_1}^{S_2}=(T_{S_2})^{-1} * T_{S_1}$. With the extrinsics between sensors, we can also leverage a few modality-specific pair-wise calibrations to form a chain of relative poses. One possible chain of extrinsic parameters is as follows:
1) GNSS-to-LiDAR/IMU extrinsic ($T_G^L$). 2) LiDAR/IMU-to-camera extrinsic ($T_L^C$).
With this chain, one can calculate any pair-wise transformation matrices. 
For instance, to find GNSS and camera correspondence, we only need to calculate $T_G^C = T_G^L * T_L^C$.



\subsection{Data Acquisition}

\noindent\textbf{Scene Configuration and Data Collection. }
By leveraging CMS platforms and the instrumented road-side unit, we can now stage diverse vehicle encountering scenarios, \eg at intersections, highway ramps, and parking lots. In these interactive scenarios, we record multimodal data, including LiDAR, images, GNSS, and C-V2X network traces with busy background traffic.
The data is collected in multiple continuous takes. Each take spans from the vehicles' departure at various starting points to their convergence at a designated encounter zone, concluding once all vehicles exit the area. To ensure scenario diversity, we vary the initial positions and driving routes across takes, while coordinating the vehicles to arrive at the target zone simultaneously.

\parab{Annotation. }
After data collection, each sensor frame is assigned a unique frame identifier based on its \textit{globally-synchronized} sensor timestamp. 
Frames across different agents are synchronized using these identifiers. In cases of frame loss, the most recent valid frame is reused to ensure that all frame sequences remain temporally consistent and of equal length.
To reduce manual annotation effort, we develop a semi-automatic labeling pipeline that integrates an object detection backbone~\cite{bevfusion}, an object tracking module~\cite{AB3DMOT}, and a late fusion module. Specifically, the detection backbone first generates coarse bounding boxes\footnote{The initial release focuses on the \textit{Car} class, future releases will expand object classes.} for each agent’s data. The tracking module then refines the bounding box positions and orientations. Then all annotations from each agent are transformed in globally consistent ENU frame, where the late fusion module is applied to refine box poses and remove redundant boxes based on the Intersection-over-Union (IoU) of detected objects. The fusion process is performed in two stages: in the first stage, overlapping boxes with high IoU values are averaged to merge their spatial information; in the second stage, overlapping boxes are further filtered by retaining only those with the highest confidence scores.
Finally, the tracking module is reapplied to refine object positions and orientations, producing preliminary labels. These preliminary labels are then manually verified and refined by trained professional annotators using~\cite{sustecth_points}, producing the ground-truth labels.

\section{\Sysname Benchmark}

In this section, we present a comprehensive benchmark evaluation on \sysname, focusing on object detection and motion prediction.
In the benchmark, the dataset is divided into train, validate, and test subsets with a ratio of 5:2:3. 
We conduct all experiments on a unified cooperative perception pipeline developed on top of MMDet3D~\cite{mmdet3d2020} and OpenCOOD~\cite{xu2022opv2v}. Our unified pipeline integrates a configurable C-V2X network simulator and provides a consistent \textit{multi-ego} training and evaluation protocol.
We retrained all perception and prediction baseline models on the \sysname dataset to allow them to adapt to the characteristics and sensor configurations, avoiding domain transfer gaps. 



\parab{Metrics.}
For perception, we adopt the mean Average Precision (mAP) 
as the primary metric, with 2D (BEV) and 3D IoU thresholds of (0.3, 0.5, 0.7). 
For motion, we use Minimum Average Displacement Error (minADE), and Minimum Final Displacement Error (minFDE) evaluated at different horizons (1s, 3s, 5s).
In our \textit{multi-ego} pipeline, we iteratively treat each of the three vehicles as the ego vehicle, with the remaining vehicles and the infrastructure as cooperators. The evaluation reports results averaged over all ego vehicles.

\parab{Baselines.}
To build a comprehensive benchmark on \sysname, we evaluate state-of-the-art  cooperative perception models, including graph-based, transformer-based fusion algorithms and recent communication-efficient methods. All models are retrained using consistent parameters, including voxelization size and detection range for fair comparison. 
Furthermore, we benchmark the cooperative motion prediction performance following the perception–prediction (P\&P) setting. The baselines include:
\begin{itemize}
    \item \textbf{V2VNet}~\cite{v2vnet}: 
    A graph-based intermediate fusion model. Vehicles share encoded features through learnable edge connections.

    \item \textbf{V2VAM}~\cite{v2vam}: 
    An attention-based transformer model, which captures intra-agent and inter-agent attention, dynamically weights agent contributions.

    \item \textbf{V2X-ViT}~\cite{v2xvit}: 
    A vision transformer-based BEV fusion architecture composed of multi-agent self-attention and multi-scale window self-attention layers.

    \item \textbf{CoBEVT}~\cite{xu2022cobevt}: 
    A Swin-Transformer-based BEV fusion architecture that applies multi-scale window attention for spatial feature alignment.

    \item \textbf{CMP}~\cite{wang2025cmp}:
    A cooperative motion prediction framework built on top of the CoBEVT perception backbone, which aggregates predictions from multiple vehicles using a transformer-based motion decoder. 
    \item \textbf{ERMVP}~\cite{ermvp}: 
    A communication-efficient and collaboration-robust model that compresses shared features via filter-and-merge sampling and calibrates pose induced misalignment with consensus sparse features.
    \item \textbf{CoSDH}~\cite{cosdh}: 
    A communication-efficient model using supply–demand-aware region selection and an intermediate–late hybrid fusion scheme to balance accuracy and bandwidth.    
\end{itemize}

\begin{table*}[t]
\centering
\renewcommand{\arraystretch}{0.75}
\caption{
Benchmark of cooperative perception under C-V2X and \textcolor{lightgray}{unlimited} (infeasible) network, with increasing number of agents, in terms of accuracy (mAP), total data sharing size per frame, and the latency if all data were to be transmitted over C-V2X.
}
\label{tab:multi_agent_extended}

\resizebox{\textwidth}{!}{
\begin{tabular}{c | c | *{3}{>{\centering\arraybackslash}p{1.2cm}} | *{3}{>{\centering\arraybackslash}p{1.2cm}} | c c}
\toprule
\textbf{Model} & \textbf{Agents} 
& \multicolumn{3}{c|}{\textbf{mAP (C-V2X)}} 
& \multicolumn{3}{c|}{\textcolor{lightgray}{\textbf{mAP (Unlimited)}}}
& \textbf{Sharing} & \textbf{Latency (ms)} \\
\cmidrule(lr){3-5}
\cmidrule(lr){6-8}
& 
& \textbf{@0.3} & \textbf{@0.5} & \textbf{@0.7}
& \textcolor{lightgray}{\textbf{@0.3}} & \textcolor{lightgray}{\textbf{@0.5}} & \textcolor{lightgray}{\textbf{@0.7}}
& \textbf{Size (MB)} & \textbf{under C-V2X } \\
\midrule
\multirow{5}{*}{V2VNet \cite{v2vnet}}
& V+I     & \textbf{0.43} & \textbf{0.33} & \textbf{0.20} & \textcolor{lightgray}{0.55} & \textcolor{lightgray}{0.42} & \textcolor{lightgray}{0.23} & 10.0 & 64,349 \\
& V+V     & 0.35 & 0.25 & 0.15 & \textcolor{lightgray}{0.66} & \textcolor{lightgray}{0.55} & \textcolor{lightgray}{0.34} & 10.0 & 64,038 \\
& V+V+I   & 0.36 & 0.27 & 0.15 & \textcolor{lightgray}{0.70} & \textcolor{lightgray}{0.60} & \textcolor{lightgray}{0.37} & 20.0 & 76,879 \\
& V+2V    & 0.32 & 0.23 & 0.13 & \textcolor{lightgray}{0.79} & \textcolor{lightgray}{0.74} & \textcolor{lightgray}{0.47} & 20.0 & 72,773 \\
& V+2V+I  & 0.32 & 0.23 & 0.13 & \textcolor{lightgray}{\textbf{0.81}} & \textcolor{lightgray}{\textbf{0.75}} & \textcolor{lightgray}{\textbf{0.50}} & 30.0 & 107,432 \\
\midrule
\multirow{5}{*}{V2X-ViT \cite{v2xvit}}
& V+I     & 0.50 & 0.38 & 0.20 & \textcolor{lightgray}{0.54} & \textcolor{lightgray}{0.41} & \textcolor{lightgray}{0.21} & 0.338 & 2,211 \\
& V+V     & 0.53 & 0.43 & 0.26 & \textcolor{lightgray}{0.65} & \textcolor{lightgray}{0.55} & \textcolor{lightgray}{0.35} & 0.338 & 1,305 \\
& V+V+I   & 0.56 & 0.45 & 0.26 & \textcolor{lightgray}{0.70} & \textcolor{lightgray}{0.59} & \textcolor{lightgray}{0.36} & 0.667 & 9,215 \\
& V+2V    & \textbf{0.60} & \textbf{0.53} & \textbf{0.33} & \textcolor{lightgray}{0.79} & \textcolor{lightgray}{0.73} & \textcolor{lightgray}{0.49} & 0.667 & 7,022 \\
& V+2V+I  & \textbf{0.60} & 0.52 & \textbf{0.33} & \textcolor{lightgray}{\textbf{0.80}} & \textcolor{lightgray}{\textbf{0.74}} & \textcolor{lightgray}{\textbf{0.50}} & 1.014 & 9,976 \\
\midrule
\multirow{5}{*}{V2VAM \cite{v2vam}}
& V+I     & 0.53 & 0.40 & 0.22 & \textcolor{lightgray}{0.57} & \textcolor{lightgray}{0.44} & \textcolor{lightgray}{0.24} & 0.423 & 2,264 \\
& V+V     & 0.53 & 0.43 & 0.26 & \textcolor{lightgray}{0.68} & \textcolor{lightgray}{0.56} & \textcolor{lightgray}{0.37} & 0.423 & 2,253 \\
& V+V+I   & 0.58 & 0.47 & 0.27 & \textcolor{lightgray}{0.74} & \textcolor{lightgray}{0.63} & \textcolor{lightgray}{0.39} & 0.846 & 10,301 \\
& V+2V    & 0.61 & \textbf{0.54} & \textbf{0.34} & \textcolor{lightgray}{0.82} & \textcolor{lightgray}{0.76} & \textcolor{lightgray}{\textbf{0.53}} & 0.846 & 8,117 \\
& V+2V+I  & \textbf{0.62} & \textbf{0.54} & \textbf{0.34} & \textcolor{lightgray}{\textbf{0.84}} & \textcolor{lightgray}{\textbf{0.78}} & \textcolor{lightgray}{\textbf{0.53}} & 1.269 & 11,302 \\
\midrule
\multirow{5}{*}{CoBEVT \cite{xu2022cobevt}}
& V+I     & 0.48 & 0.39 & 0.20 & \textcolor{lightgray}{0.52} & \textcolor{lightgray}{0.43} & \textcolor{lightgray}{0.21} & 0.500 & 2,707 \\
& V+V     & 0.50 & 0.41 & 0.26 & \textcolor{lightgray}{0.64} & \textcolor{lightgray}{0.55} & \textcolor{lightgray}{0.37} & 0.500 & 2,393 \\
& V+V+I   & 0.54 & 0.44 & 0.25 & \textcolor{lightgray}{0.70} & \textcolor{lightgray}{0.60} & \textcolor{lightgray}{0.37} & 1.000 & 11,059 \\
& V+2V    & 0.57 & 0.49 & \textbf{0.33} & \textcolor{lightgray}{0.80} & \textcolor{lightgray}{0.73} & \textcolor{lightgray}{\textbf{0.54}} & 1.000 & 8,987 \\
& V+2V+I  & \textbf{0.59} & \textbf{0.50} & 0.32 & \textcolor{lightgray}{\textbf{0.82}} & \textcolor{lightgray}{\textbf{0.75}} & \textcolor{lightgray}{\textbf{0.54}} & 1.500 & 12,381 \\
\midrule
\multirow{5}{*}{ERMVP \cite{ermvp}}
& V+I     & 0.61 & 0.46 & 0.24 & \textcolor{lightgray}{0.65} & \textcolor{lightgray}{0.51} & \textcolor{lightgray}{0.26} & 0.346 & 1,730 \\
& V+V     & 0.61 & 0.49 & 0.30 & \textcolor{lightgray}{0.76} & \textcolor{lightgray}{0.63} & \textcolor{lightgray}{0.42} & 0.346 & 1,730 \\
& V+V+I   & 0.63 & 0.51 & 0.31 & \textcolor{lightgray}{0.79} & \textcolor{lightgray}{0.67} & \textcolor{lightgray}{0.43} & 0.692 & 3,460 \\
& V+2V    & \textbf{0.66} & \textbf{0.57} & \textbf{0.37} & \textcolor{lightgray}{\textbf{0.87}} & \textcolor{lightgray}{0.80} & \textcolor{lightgray}{\textbf{0.57}} & 0.692 & 3,460 \\
& V+2V+I  & 0.64 & 0.56 & 0.36 & \textcolor{lightgray}{\textbf{0.87}} & \textcolor{lightgray}{\textbf{0.81}} & \textcolor{lightgray}{\textbf{0.57}} & 1.038 & 5,190 \\
\midrule
\multirow{5}{*}{CoSDH \cite{cosdh}}
& V+I     & 0.57 & 0.43 & 0.23 & \textcolor{lightgray}{0.58} & \textcolor{lightgray}{0.44} & \textcolor{lightgray}{0.24} & 0.0595 & 298 \\
& V+V     & 0.63 & 0.47 & 0.29 & \textcolor{lightgray}{0.67} & \textcolor{lightgray}{0.54} & \textcolor{lightgray}{0.35} & 0.0595 & 298 \\
& V+V+I   & 0.68 & 0.53 & 0.32 & \textcolor{lightgray}{0.72} & \textcolor{lightgray}{0.62} & \textcolor{lightgray}{0.39} & 0.1190 & 595 \\
& V+2V    & 0.74 & \textbf{0.61} & 0.40 & \textcolor{lightgray}{0.79} & \textcolor{lightgray}{0.72} & \textcolor{lightgray}{0.51} & 0.1190 & 595 \\
& V+2V+I  & \textbf{0.75} & \textbf{0.61} & \textbf{0.41} & \textcolor{lightgray}{\textbf{0.81}} & \textcolor{lightgray}{\textbf{0.73}} & \textcolor{lightgray}{\textbf{0.52}} & 0.1785 & 893 \\

\bottomrule

\end{tabular}
}
\vspace{-3mm}
\end{table*}

\parab{Detection Benchmark Results.} \tabref{tab:multi_agent_extended} summarizes the perception benchmark. We benchmark each baseline over five cooperative settings; \textbf{1) V+I}: with only infra, \textbf{2) V+V}: with only one peer vehicle, \textbf{3) V+V+I}: with one vehicle and infra, \textbf{4) V+2V}: with two peer vehicles, and \textbf{5) V+2V+I}: with all three cooperative agents. 
In addition to showing the mAP under C-V2X bandwidth constraints, we also benchmark the performance with an unlimited network assumption (in \textcolor{lightgray}{lightgray} color) as done in other existing benchmarks~\cite{xu2023v2v4real, tumtraf}, to show the original benefits of fusing rich features from multiple vantage points. 

Overall, all cooperative perception models benefit from information sharing among vehicles and infrastructure. As the number of cooperative agents grows, \textit{almost all} (except V2VNet) methods achieve increasingly better detection performance than non-cooperative settings (\tabref{tab:single_agent}), under both C-V2X and unlimited settings, across different IoU thresholds. 
However, 
the mAPs under realistic C-V2X bandwidth are upto 49\% lower than the grayed-out numbers under an unlimited network. This occurs because the required data transmission size exceeds C-V2X network capabilities, which results in late data delivery. 
The drastic difference shows the gap between idealized perception and bandwidth-constrained real-world deployment.

Secondly, for \textit{non-communication-efficient} methods (the top four), due to the mismatch between the sharing volume and the limited C-V2X network capacity, cooperation can only happen \textit{very infrequently}, ranging from every 2 to 100s of seconds, while hundreds of frames in between the interval can not benefit from cooperation. 
Consequently, agents receive \textit{outdated, stale} data, which fails to improve, but can actively harm, the perception accuracy. 
Furthermore, adding more participants only exacerbates channel contention and increases latency. Hence, the benefit of an additional vantage point may effectively degrade end-to-end performance due to the increased staleness (\eg for V2VNet, and CoBEVT at V+2V+I, where sharing size is large). 
In contrast, the communication-aware baselines ERMVP~\cite{ermvp} and CoSDH~\cite{cosdh} suffer less from this contention. They transmit only the most informative features within a bandwidth budget, substantially reducing the per-frame sharing size. While they achieve the strongest deployable accuracy under C-V2X among all baselines, the latency does not scale gracefully and is still too long for vehicles to make safety-critical reactions.
This result corroborates the value of bandwidth reduction research~\cite{hu2022where2comm, CooperTrim} in cooperative perception, motivating future research to further resolve the \textit{mismatch} between sharing size and network capacity.


\begin{table}[t]
\centering

\begin{minipage}[t]{0.48\textwidth}
\centering
\renewcommand{\arraystretch}{0.75}
\caption{Single agent perception performance using different backbones.}
\label{tab:single_agent}

\footnotesize
\setlength{\tabcolsep}{4pt}

\resizebox{\textwidth}{!}{
\begin{tabular}{l l c c c}
\toprule
\textbf{Agent} & \textbf{Method} & \textbf{mAP@0.3} & \textbf{mAP@0.5} & \textbf{mAP@0.7} \\
\midrule

\multirow{2}{*}{Infra}
& PointPillars~\cite{pointpillars} & \textbf{0.24} & 0.09 & 0.04 \\
& SECOND~\cite{yan2018second} & 0.20 & \textbf{0.11} & \textbf{0.06} \\

\midrule
\multirow{2}{*}{CAV A}
& PointPillars~\cite{pointpillars} & \textbf{0.37} & 0.20 & 0.08 \\
& SECOND~\cite{yan2018second} & 0.36 & \textbf{0.23} & \textbf{0.11} \\

\midrule
\multirow{2}{*}{CAV B}
& PointPillars~\cite{pointpillars} & \textbf{0.38} & 0.20 & 0.08 \\
& SECOND~\cite{yan2018second} & 0.35 & \textbf{0.24} & \textbf{0.13} \\

\midrule
\multirow{2}{*}{CAV C}
& PointPillars~\cite{pointpillars} & \textbf{0.34} & 0.18 & 0.07 \\
& SECOND~\cite{yan2018second} & \textbf{0.34} & \textbf{0.22} & \textbf{0.11} \\

\bottomrule
\end{tabular}
}
\end{minipage}
\hfill
\begin{minipage}[t]{0.48\textwidth}
\centering
\renewcommand{\arraystretch}{0.9}
\caption{
Multi-modal multi-agent cooperative perception benchmark.
}
\label{tab:multi_modality_eval}

\footnotesize
\setlength{\tabcolsep}{3pt}

\resizebox{\textwidth}{!}{
\begin{tabular}{l l c c c c c c}
\toprule
\textbf{Agent} & \textbf{Modality} &
\multicolumn{3}{c}{\textbf{mAP(BEV)}} &
\multicolumn{3}{c}{\textbf{mAP(3D)}} \\
\cmidrule(lr){3-5}
\cmidrule(lr){6-8}
& & \textbf{@0.3} & \textbf{@0.5} & \textbf{@0.7}
& \textbf{@0.3} & \textbf{@0.5} & \textbf{@0.7} \\
\midrule

\multirow{2}{*}{V}
& LiDAR & 0.78 & 0.56 & 0.31 & 0.70 & 0.38 & 0.21 \\
& LiDAR+Cam & \textbf{0.80} & \textbf{0.57} & \textbf{0.33} & \textbf{0.72} & \textbf{0.42} & \textbf{0.23} \\
\midrule
\multirow{2}{*}{V+V}
& LiDAR & \textbf{0.81} & \textbf{0.65} & 0.39 & \textbf{0.75} & \textbf{0.54} & 0.22 \\
& LiDAR+Cam & 0.80 & \textbf{0.65} & \textbf{0.41} & 0.74 & \textbf{0.54} & \textbf{0.25} \\
\midrule
\multirow{2}{*}{V+2V}
& LiDAR & \textbf{0.90} & \textbf{0.84} & 0.56 & \textbf{0.87} & 0.73 & 0.30 \\
& LiDAR+Cam & \textbf{0.90} & 0.83 & \textbf{0.61} & \textbf{0.87} & \textbf{0.74} & \textbf{0.35} \\

\bottomrule
\end{tabular}
}

\end{minipage}

\end{table}

\parab{Multi-modal Cooperative Perception.}
To benchmark multi-modal cooperative perception, we developed a multi-modal cooperative framework based on top of BEVFusion~\cite{bevfusion}.
\tabref{tab:multi_modality_eval} compares the performance under LiDAR-only and LiDAR+Camera configurations.
The results show that when cooperative perception is combined with multi-modal sensing, the performance gain becomes more significant, especially at \textit{higher IoU thresholds}, where the model benefits more from the complementary modality. This indicates that both inter-agent collaboration and multi-modal feature fusion contribute to perception understanding. 

\parab{Motion Prediction Benchmark Results.}
To further evaluate the capability of \sysname in supporting downstream tasks, we benchmark the cooperative motion prediction performance of V2VNet~\cite{v2vnet} and CMP~\cite{wang2025cmp} following the perception–prediction (P\&P) setting. In this setting, cooperative perception first aggregates multi-agent features, which are then used by the motion prediction model to predict future trajectories.
In the P\&P setting, the communication cost is dominated by the perception feature exchange, while the data exchange in the subsequent prediction module (\eg in CMP) is negligible. Therefore, we omit the communication characteristics, and refer to those reported in \tabref{tab:multi_agent_extended} under different settings. The quantitative results are summarized in \tabref{tab:motion_prediction}. Overall, cooperative perception improves motion prediction accuracy, especially for \textit{longer prediction horizons} for both models. 
Between the two models, CMP outperforms V2VNet across all evaluation horizons. This performance gap highlights the advantage of a transformer-based motion decoder in modeling complex multi-agent interactions. V2VNet's short-horizon results are interesting, as it is counterintuitive. We can only suspect that the additional information from peer vehicles disturbs the intermediate fusion for short-horizon predictions. We believe there is still a huge research space into cooperative motion prediction, as well as other downstream tasks beyond detection.

\begin{table*}[t]
\centering
\caption{\sysname benchmark on cooperative motion prediction}
\label{tab:motion_prediction}

\footnotesize
\resizebox{\textwidth}{!}{
\begin{tabular}{l|l|ccc|ccc}
\toprule
\textbf{Model} & \textbf{Agent}
& \textbf{minADE@1s}$\downarrow$ & \textbf{minADE@3s}$\downarrow$ & \textbf{minADE@5s}$\downarrow$
& \textbf{minFDE@1s}$\downarrow$ & \textbf{minFDE@3s}$\downarrow$ & \textbf{minFDE@5s}$\downarrow$ \\
\midrule

\multirow{3}{*}{V2VNet~\cite{v2vnet}}
& V     & \textbf{0.4895} & 2.1648 & 4.4538 & \textbf{0.9307} & 5.7344 & 10.6390 \\
& V+V   & 0.5424 & 1.5611 & 3.5591 & 0.7528 & 4.4820 & 9.1667 \\
& V+2V  & 0.7610 & \textbf{1.3796} & \textbf{2.9096} & 1.0079 & \textbf{3.4762} & \textbf{7.3660} \\
\midrule

\multirow{3}{*}{CMP~\cite{wang2025cmp}}
& V     & 0.3851 & 0.8421 & 1.4022 & 0.5786 & 1.6125 & 3.0712 \\
& V+V   & 0.4076 & 0.9330 & 1.6416 & 0.6143 & 1.8492 & 3.7389 \\
& V+2V  & \textbf{0.3214} & \textbf{0.7252} & \textbf{1.2893} & \textbf{0.4723} & \textbf{1.4551} & \textbf{2.9716} \\

\bottomrule
\end{tabular}
}
\vspace{-2mm}
\end{table*}


\begin{figure}[t]
\centering
\begin{minipage}{0.15\columnwidth}
\vspace{-2mm}
\includegraphics[width=\textwidth]{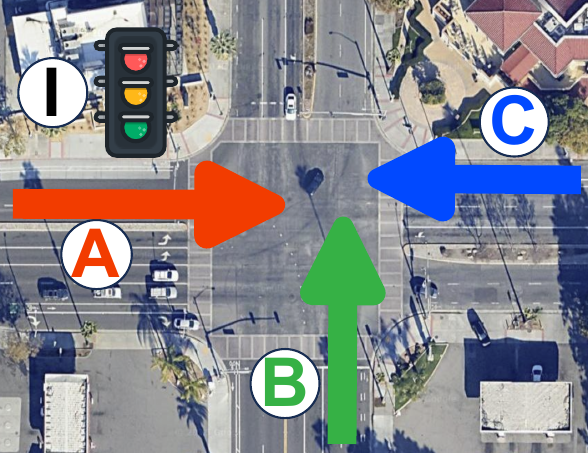}
\end{minipage}
\begin{minipage}{0.84\columnwidth}
\includegraphics[width=\textwidth]{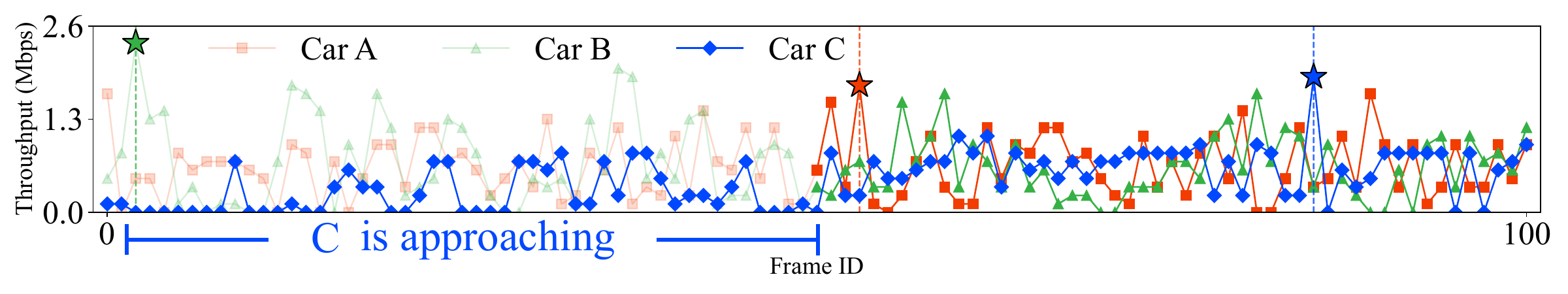}
\end{minipage}
\caption{C-V2X Communication throughput variation across different frames. Pentagram denotes the peak throughput.}
\label{fig:tput_per_frame}
\vspace{-6mm}
\end{figure}

\parab{C-V2X Network Performance Analysis.}
\figref{fig:tput_per_frame} shows the variation of throughput across frames. The C-V2X network exhibits highly dynamic behavior due to vehicle mobility and channel contention among multiple agents. As a result, no single agent can occupy the channel continuously, and the effective throughput fluctuates significantly over time. These observations highlight the need for careful spectrum sharing design and motivate the use of more adaptive cooperation strategies~\cite{CooperTrim}.
In addition, we empirically evaluated the feasibility of transmitting cooperative perception data using the baselines. The data size and corresponding transmission time are summarized in \tabref{tab:multi_agent_extended}. The data were segmented into 2100 bytes chunks and transmitted sequentially over the C-V2X radio, with each packet sent at a 1~ms interval\footnote{Corresponding to the minimum time resource in C-V2X standard~\cite{3gpp36213}}. All models require more than 100 ms to deliver a single frame, resulting in outdated data and providing no significant improvement in perception performance.

\begin{wrapfigure}{r}{0.6\textwidth}
\vspace{-10mm} 
\centering
\renewcommand{\arraystretch}{0.85}
\setlength{\tabcolsep}{3.2pt}
\captionof{table}{Evaluation under different network settings}
\label{tab:rebuttal_v2x}
\resizebox{\linewidth}{!}{
\begin{tabular}{lcccccc}
\toprule
\textbf{Method} & \textbf{AP$_{30}$} & \textbf{AP$_{50}$} & \textbf{AP$_{70}$} & \textbf{Pkt. Loss} & \textbf{Latency} & \textbf{Tput} \\
&  &  &  & \textbf{(\%)} & \textbf{(ms)} & \textbf{(Mbps)} \\
\midrule
Local      & 0.18 & 0.15 & 0.04 & --    & --    & --   \\
C-V2X      & 0.21 & 0.19 & 0.05 & 38.19 & 31.46 & 0.72 \\
Unlimited  & 0.24 & 0.21 & 0.06 & 0     & $<$1  & 1.25 \\
\bottomrule
\vspace{3mm}
\end{tabular}
}

\vfill

\includegraphics[width=0.49\linewidth]{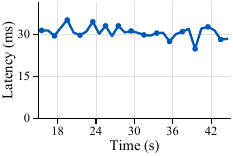}
\hfill
\includegraphics[width=0.49\linewidth]{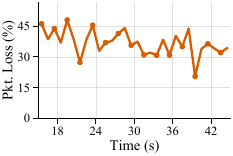}
\vspace{-6mm}
\captionof{figure}{Latency and packet loss rate under C-V2X}
\label{fig:rebuttal_v2x}
\vspace{-6mm}
\end{wrapfigure}

While bandwidth is the dominant bottleneck, other communication characteristics also affect end-to-end performance. To isolate their impact, we remove the bottleneck by conducting an oracle-style experiment: each CAV transmits only the down-sampled LiDAR points bounded by ground-truth boxes, reducing the data volume below the C-V2X capacity. As shown in \tabref{tab:rebuttal_v2x}, even when bandwidth is sufficient in both C-V2X and unlimited, packet loss and latency still degrade detection accuracy relative to the unlimited setting. \figref{fig:rebuttal_v2x} further shows that such packet loss and latency fluctuate over time and are inevitable in real-world channels due to vehicle mobility, interference, and channel contention. This indicates that resolving the bandwidth gap alone is insufficient; robustness to packet loss and latency is equally important for deployable cooperative perception.

Overall, the results reveal a substantial gap between the assumptions made in existing works and the actual operational characteristics of real-world C-V2X networks, underscoring the importance of network-aware cooperative perception design, which is a central goal of \sysname.

\section{Data Quality Validation}
\begin{figure}
\centering
\vspace{-3mm}
\subfloat[Intra-agent (LiDAR-Camera)]{
    \includegraphics[width=0.47\linewidth]{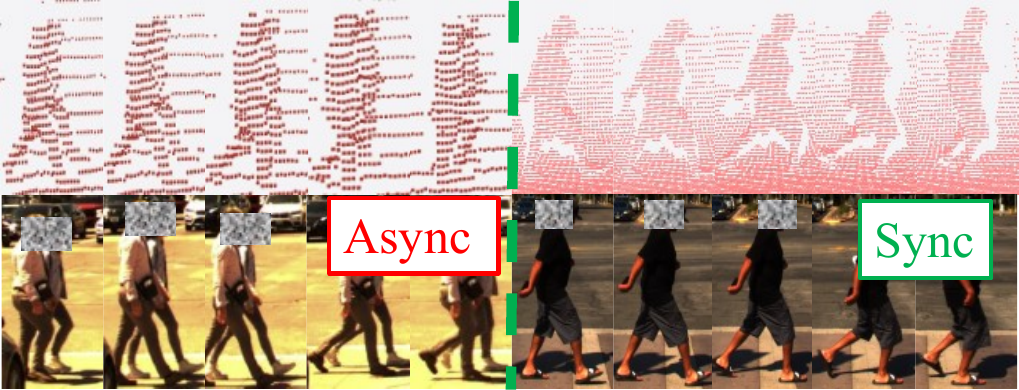}
    \label{fig:lidar_camera_seq}
}\hfill
\subfloat[Inter-agent (Multi-platform)]{
    \includegraphics[width=0.47\linewidth]{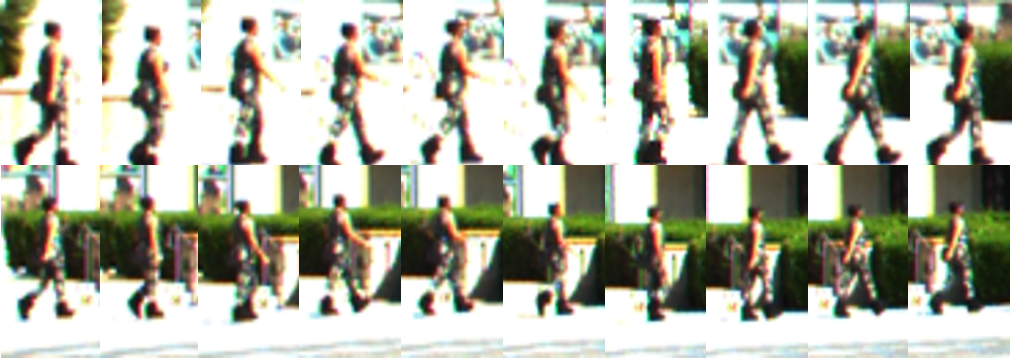}
    \label{fig:multi_camera_seq}
}

\caption{Visualization of synchronized sequences in \sysname.}
\label{fig:synchronized_sequence}
\vspace{-6mm}
\end{figure}

\parab{Synchronization.}
\figref{fig:synchronized_sequence} presents an example synchronization sequence from the \sysname dataset. \figref{fig:lidar_camera_seq} shows onboard LiDAR and camera sequences capturing pedestrians crossing the intersection before and after synchronization. With PTP and hardware triggering synchronization, the LiDAR and camera captures the motion of the pedestrian at the same time compared to asynchronous data. \figref{fig:multi_camera_seq} displays camera images from multiple vehicles capturing a pedestrian walking along the pathway, demonstrating consistent motion across agents and confirming effective cross-platform synchronization.

\begin{wrapfigure}{r}{0.6\columnwidth}
\centering
\vspace{-6mm}
\subfloat[RMSE PDF]{
\centering
    \includegraphics[width=0.27\columnwidth]{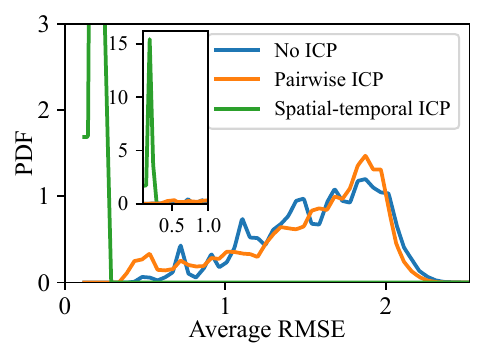}
    \label{fig:icp_diff}
}
\subfloat[RMSE comparison]{
\centering
    \includegraphics[width=0.27\columnwidth]{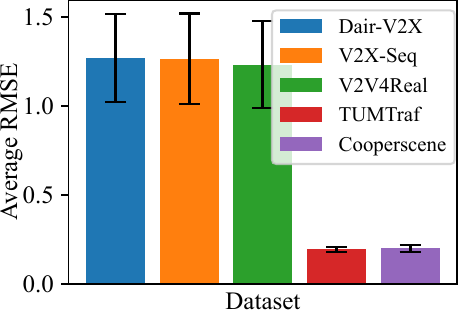}
    \label{fig:icp_comparison}
}

\caption{Spatial alignment error comparison between different methods~(a), and among other datasets (b).}
\label{fig:icp_comparsion_big}

\vspace{-6mm}
\end{wrapfigure}
\parab{Spatial Alignment.}
We quantify spatial alignment accuracy using the root mean square error (RMSE) of overlapping areas between synchronized LiDAR frames for each agent pair. Specifically, we run ICP once more on the already aligned frames, and compute the RMSE of the Euclidean distances between the matched point correspondences, measuring the misalignment between agents after convergence.
\figref{fig:icp_diff} shows the RMSE of \sysname under different spatial alignment methods. No ICP indicates that LiDAR frames are projected into the ENU coordinate system described in \secref{sec:icp} without further correction. Pairwise ICP aligns each vehicle to the infrastructure after ENU projection. Spatial-temporal ICP, which considers all agents jointly, substantially improves alignment accuracy, achieving an RMSE of 0.2~m for most frames.
\figref{fig:icp_comparison} compares the RMSE of \sysname with other real-world datasets. \sysname achieves the lowest average RMSE among all datasets except TumTraf. TumTraf aligns a single vehicle, whereas \sysname must jointly align multiple agents. Furthermore, TumTraf applies ICP every 10 frames, while \sysname achieves comparable performance with difference in only 0.01~$m$ by applying ICP on just a single frame per vehicle, highlighting the efficiency and robustness of our spatial alignment.

\begin{figure}[t]
    \centering
    \includegraphics[width=0.245\textwidth]{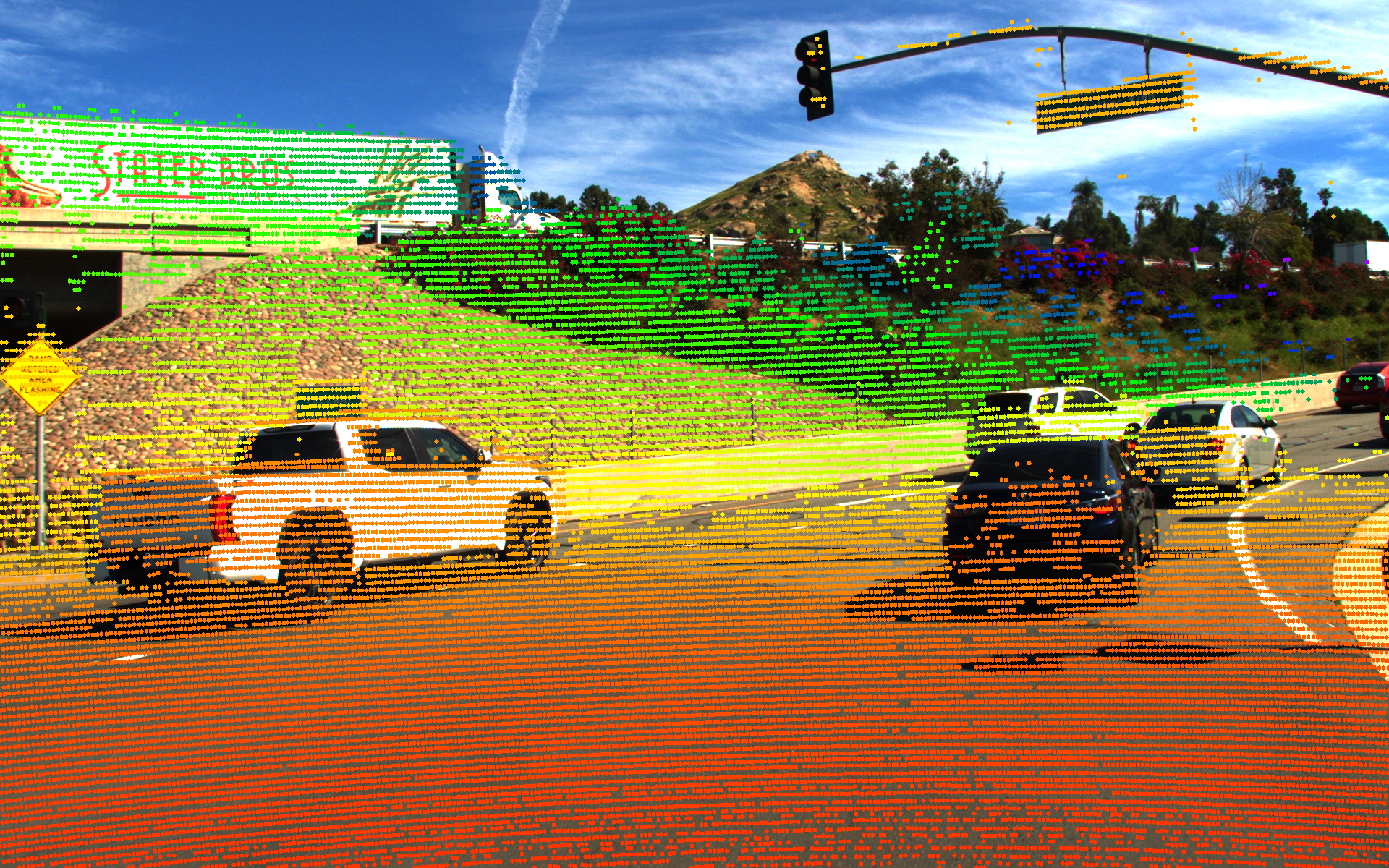}\hfill
    \includegraphics[width=0.245\textwidth]{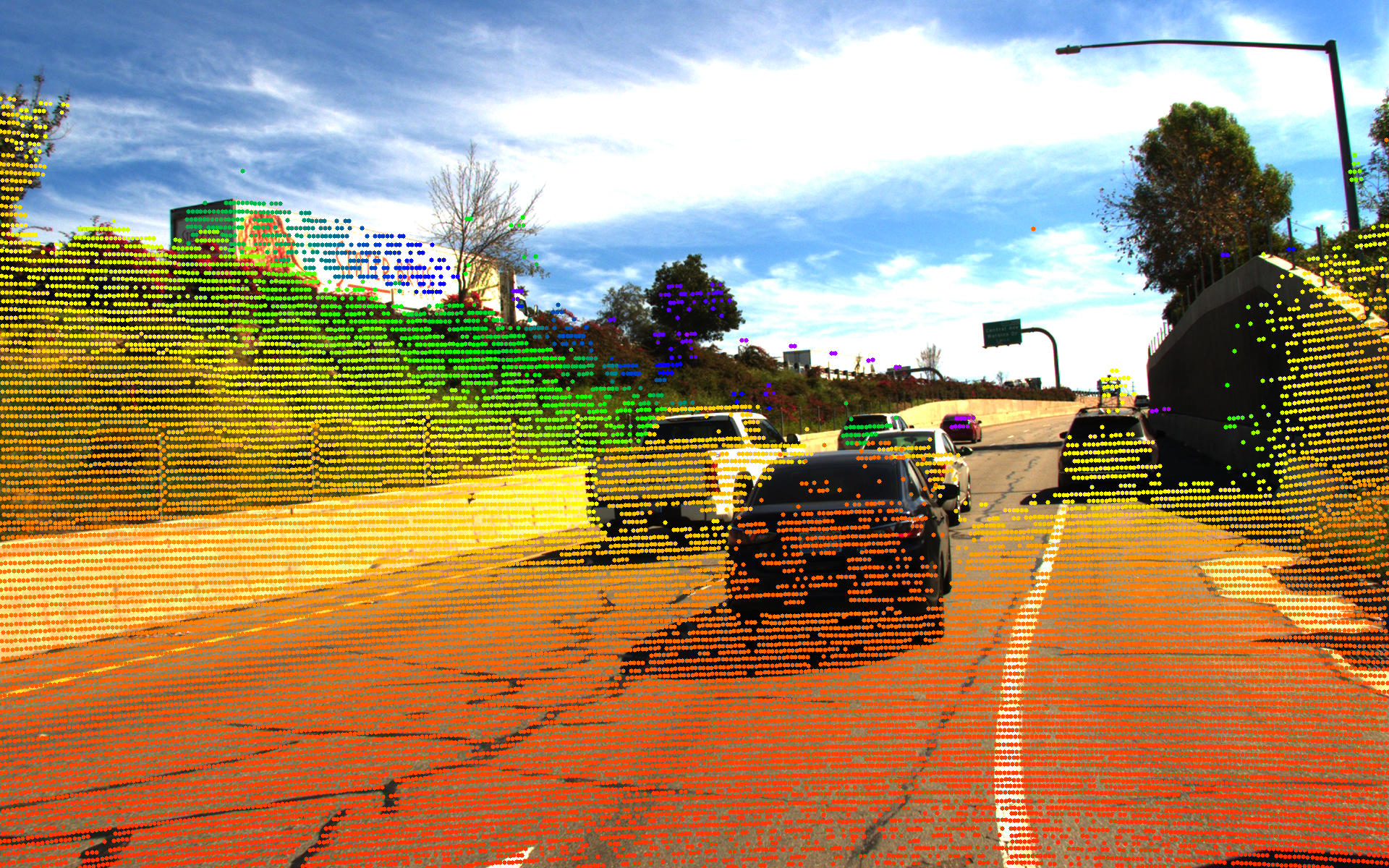}\hfill
    \includegraphics[width=0.245\textwidth]{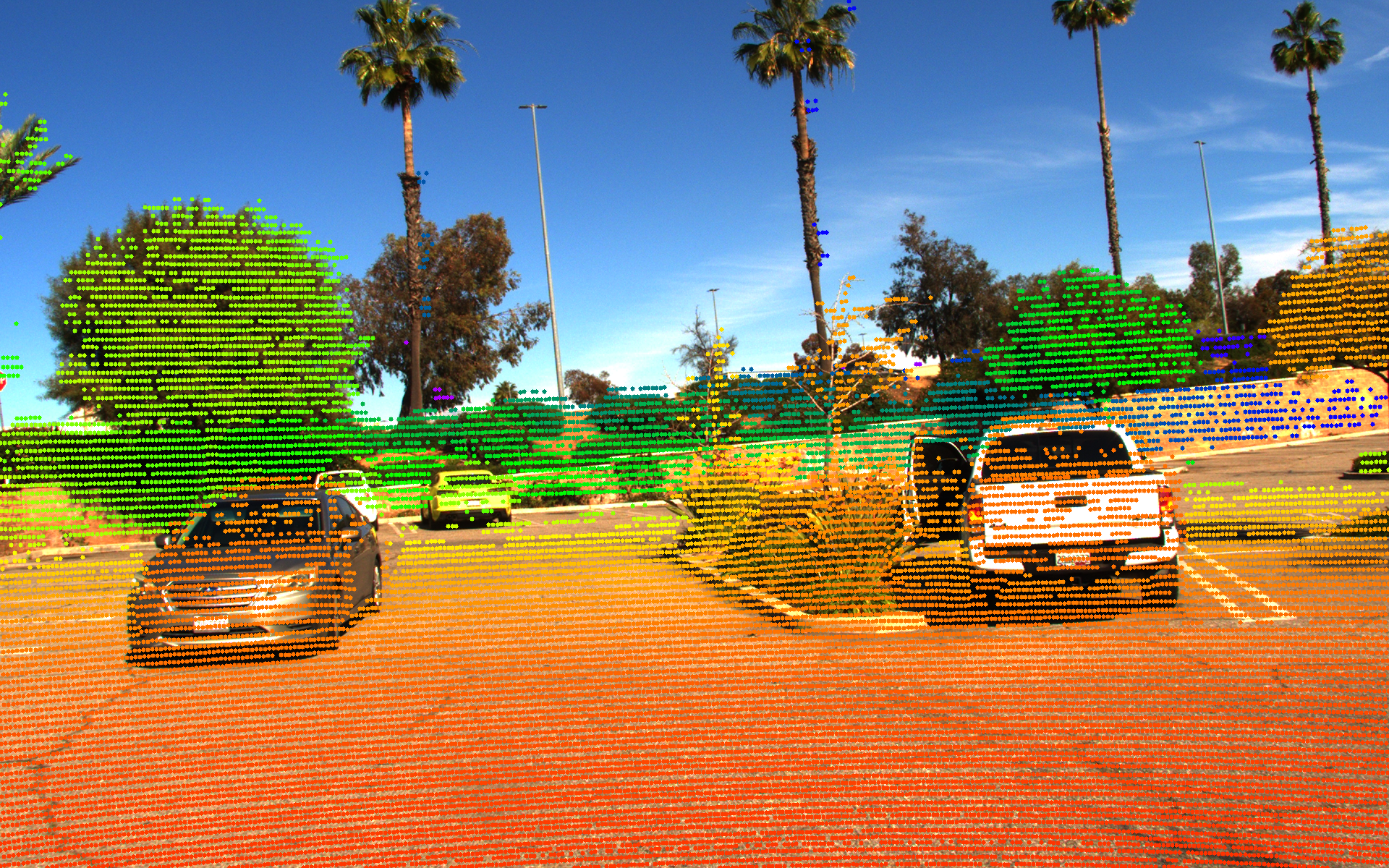}\hfill
    \includegraphics[width=0.245\textwidth]{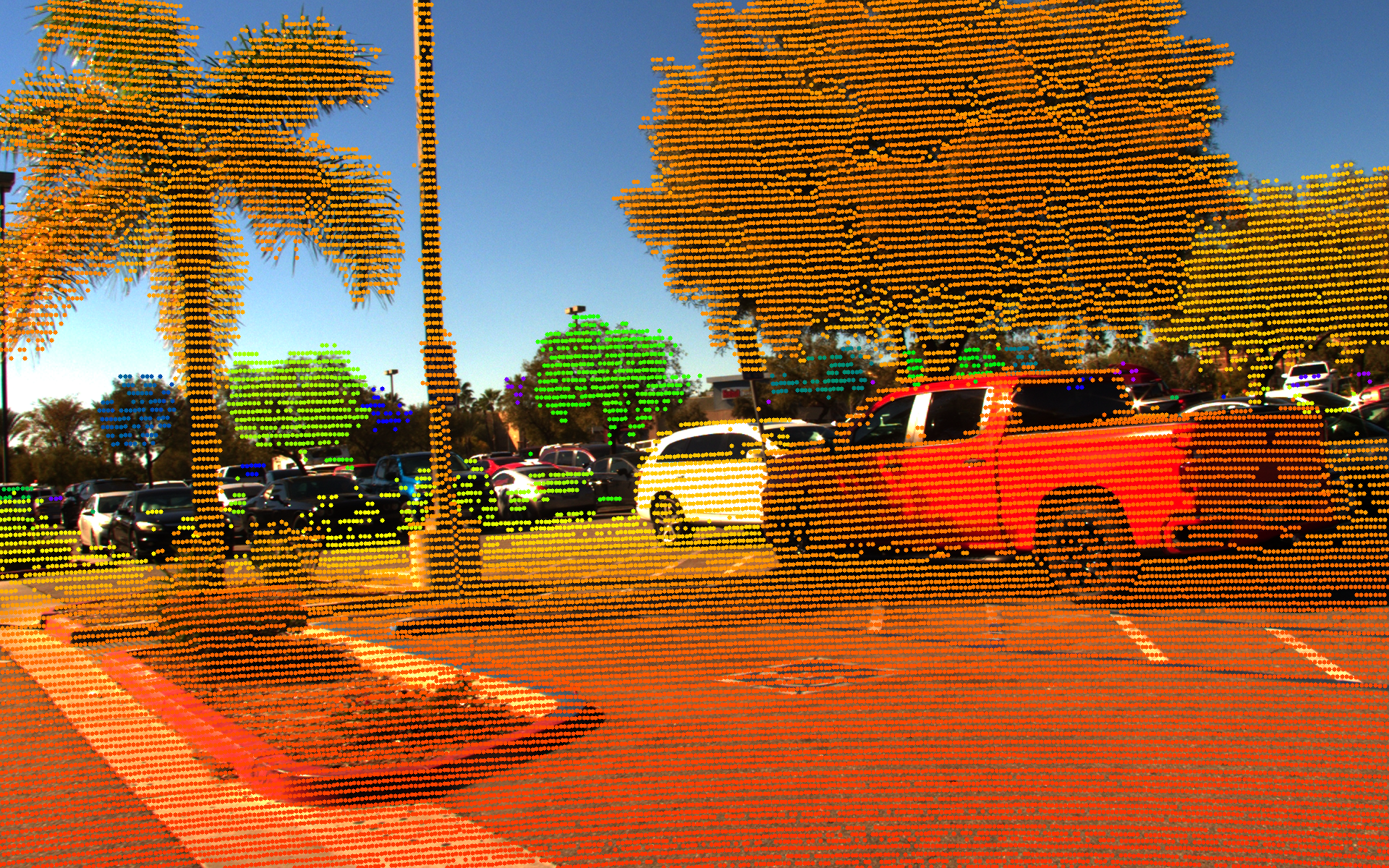}
    \caption{LiDAR-to-camera overlays using calibrated extrinsic.}
    \label{fig:outdoor_overlays_refined}
    \vspace{-6mm}
\end{figure}

\begin{wrapfigure}{r}{0.5\columnwidth}
\vspace{-8mm}
\includegraphics[width=\linewidth]{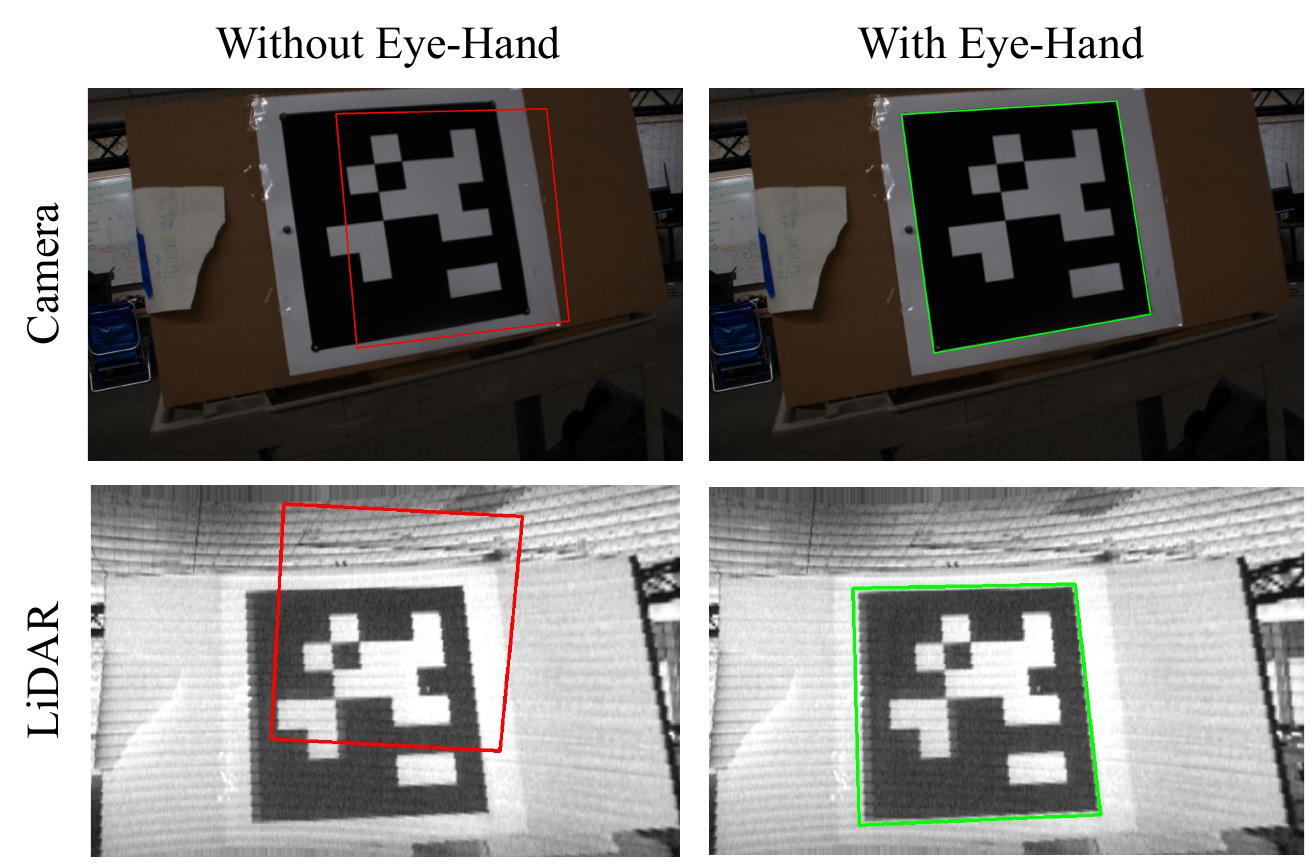}
\caption{Reprojection Qualitative Analysis}
\label{fig:reprojection}
\vspace{-6mm}
\end{wrapfigure}
\parab{Calibration.} \figref{fig:outdoor_overlays_refined} shows the results of the extrinsic calibration as LiDAR-Camera overlays, where the LiDAR's point cloud aligns with objects such as buildings, cars, trees, and poles in the image. 
We provide a qualitative analysis of our calibration by confirming accurate extrinsics through reprojection analysis. Specifically, we use the fiducial tag's pose $\mathcal{T}_{tag}^{mocap}$ and the sensor's pose $\mathcal{T}_{sensor}^{mocap}$ in the MoCap frame. By transforming and projecting the fiducial tag's corners into a sensor's frame with  
$\mathcal{T}_{tag}^{sensor}  = (\mathcal{T}_{sensor}^{mocap})^{-1} * \mathcal{T}_{tag}^{mocap}$), we can then project the corners into a sensor's image as shown by \figref{fig:reprojection}. For both sensors, the reprojection error is minimized after applying eye-hand calibration.

\section{Conclusion}

In this paper, we introduce \sysname, a real-world multi-agent, multi-modal cooperative autonomy dataset with synchronized real-world C-V2X communication characterization. 
\Sysname provides tightly aligned multi-modal multi-agent data via high precision localization, synchronization, and calibration. The benchmark
enables the research community to jointly evaluate perception performance and the practical constraints imposed by real-world vehicular networks. Evaluation results highlight a key insight: the bandwidth demands of state-of-the-art cooperative perception models far exceed the highly variable and limited throughput achievable by current C-V2X. This gap emphasizes the need for communication-efficient, network-aware cooperative perception approaches.
\sysname bridges this gap, provides a unified dataset to  
advance cooperative autonomy toward scalable, reliable, and deployable real-world systems.

\newpage
\noindent\parab{Acknowledgement.}
This work was conducted at the Collaborative Intelligence Systems Lab (CISL) at UC Riverside. CISL research is supported in part by NSF (ITE-2453817), OUSD (R\&E)/RT\&L (Center of Excellence W911NF-20-2-0267), and USDOT (Carnation 69A3552348324, NCST 69A3552348319). The views and conclusions contained in this document are those of the authors and should not be interpreted as representing the official policies, either expressed or implied, of the funding agencies.
We sincerely thank the college staff led by Victor Hill, as well as the university library staff led by Barbara Neda, for their invaluable support in data maintenance and release. We also express our gratitude to Haoge Zhou, Zhaoze Sun, Chuheng Wei, Ayoub Elazami Elidrissi, Dylon Wong, Joshua Ha, Shubham Derhgawen, Anatulya Nandi, Jonathan Setiabudi, Athena Nelson, Shashwat Jha, Jianpeng Yao, Terek Johnson, Rakshith Mahishi, Divyank Shah, Bhargav HS, Joong Ho Kim, Mason Audet, Shilpa Mukhopadhyay, Xuanpeng Zhao for their effort and participation in data collection and annotation.

{\small
\bibliographystyle{splncs04}
\bibliography{main,Q}
}


\newpage
\section*{Appendix}
\appendix
\renewcommand{\thesection}{\Alph{section}}

\section{Limitations and Discussions}
\Sysname presents a significant step in real-world cooperative perception, though we acknowledge specific limitations that outline the direction for future development. First, regarding annotation scope, the current release prioritizes the \textit{Car} class, which dominates traffic volume in our scenarios. We plan to extend annotations to more classes, such as pedestrians and cyclists, in future updates to support safety-critical research.
Second, while we provide fundamental C-V2X network traces, the dataset does not yet capture lower-layer information such as wireless resources scheduling or receiving signal strengths. Integrating these granular metrics remains a priority to enable deeper research into cross-layer design. Finally, CMS tower in \sysname currently only equipped with one single front-facing vehicle camera and LiDAR-only infrastructure. We will include more viewpoints in the future. 

\section{Implementation Detail}
\subsection{Hardware Setup}
\begin{table*}
\vspace{-5mm}
\caption{Sensor specifications.}
\label{tab:sensor_spec}
\begin{tabular}{|l|l|}
\hline
\textbf{Sensor} & \textbf{Details} \\ \hline
Vehicle LiDAR & Ouster OS2-128 REV7, 200~$m$ range, \ang{360}$\times$\ang{22.5} FoV (H $\times$ V)~\cite{Ouster} \\ \hline
Infrastructure LiDAR & Ouster OS2-64 REV7, 200~$m$ range, \ang{360}$\times$\ang{22.5} FoV (H $\times$ V)~\cite{Ouster}\\ \hline
Vehicle Camera & Lucid TRI54S, $2880 \times 1860$ resolution, 20.8 FPS~\cite{Lucid} \\ \hline
IMU & Xsens MTi-680, \ang{0.2} RMS (Roll/Pitch), \ang{0.5} RMS (Yaw)~\cite{xsense} \\ \hline
GNSS Antenna & Tallysman TW8889, Dual Band~\cite{gnss_antenna} \\ \hline
C-V2X Radio & Cohda MK6, LTE-V2X Release 14~\cite{Cohda} \\ \hline
\end{tabular}
\vspace{-10mm}
\end{table*}
\begin{table}[]
\centering
\caption{C-V2X Radio configuration.}
\label{tab:radio_params}
\begin{tabular}{|l|l|}
\hline
\textbf{Parameter}          & \textbf{Value} \\ \hline
Bandwidth                   & 20 MHz         \\ \hline
Modulation and coding scheme & Automatic~\cite{SAEJ3161}  \\ \hline
Transmission mode           & Eventflow mode~\cite{SAEJ3161} \\ \hline 
Transmitting power          & 20 dBm    \\ \hline
Enable retransmission            & True         \\ \hline
Priority                    & 3              \\ \hline
\end{tabular}
\end{table}
\tabref{tab:sensor_spec} summarizes the sensor specifications used in \sysname. As C-V2X radio parameters significantly influence communication performance~\cite{seev2x}, we detail our specific configuration in \tabref{tab:radio_params}. Each MK6 is equipped with two 5dBi antennas. Additional hardware includes a MikroTik PoE switch~\cite{switch} and a Thunderbolt Ethernet adapter~\cite{adaptor}.

\subsection{Experiment Setup}

\parab{PTP setup.} \Sysname employs 

\begin{wrapfigure}{r}{0.52\columnwidth}
\vspace{-50pt}
\centering
\includegraphics[width=0.50\columnwidth]{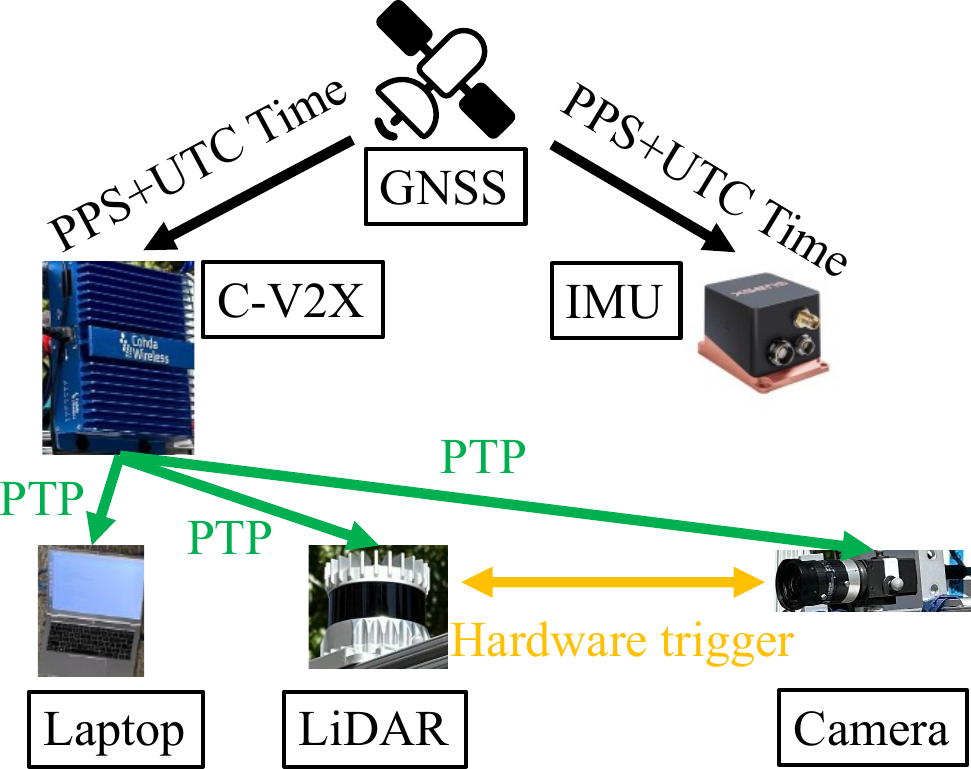}
\vspace{-6pt}
\caption{Synchronization structure of \sysname.}
\label{fig:clock_tree}
\vspace{-20pt}
\end{wrapfigure} PTP (ptp4l v1.6~\cite{ptp4l}) to synchronize all sensors, excluding the IMU. The Xsens IMU synchronizes directly via PPS signals and UTC time from GNSS. All other sensors are disciplined to the MK6, which acts as the PTP grandmaster. The synchronization structure of \sysname is illustrated in \figref{fig:clock_tree}.


\begin{wrapfigure}{r}{0.52\columnwidth}
\vspace{-15pt}
\centering
\includegraphics[width=0.50\columnwidth]{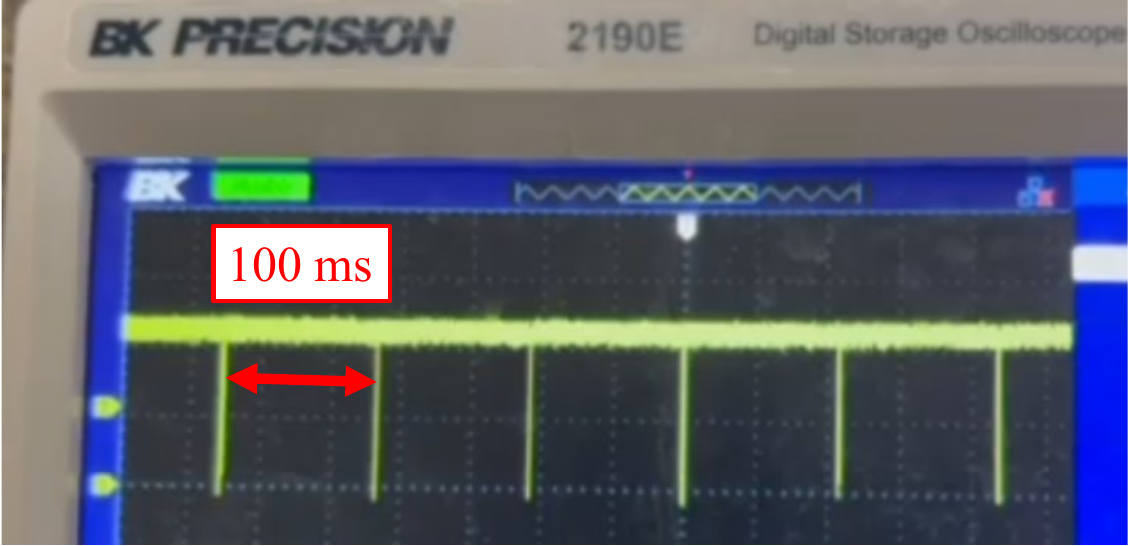}
\vspace{-6pt}
\caption{Visualization of the LiDAR trigger output signal displayed on a 100~MHz oscilloscope.}
\label{fig:lidar_freq}
\vspace{-30pt}
\end{wrapfigure}
\parab{Hardware trigger.}
To enable hardware triggering, we configure the LiDAR to output a consistent $10$~Hz pulse via GPIO as shown in \figref{fig:lidar_freq}, which serves as the capture trigger for the camera. We modified the camera driver to support PTP synchronization and hardware triggering simultaneously, ensuring LiDAR frames and camera images share identical timestamps and temporal alignment.

\parab{C-V2X measurement setup.}
We evaluate C-V2X network performance using Iperf2~\cite{iperf}. Each agent executes one Iperf client for packet transmission and three Iperf servers to receive traffic from peer agents simultaneously. Clients are configured for UDP transmission at a rate of $5$~Mbps, with servers reporting metrics at $100$~ms intervals.

\subsection{Training Detail}
\Sysname benchmark follows the OpenCOOD~\cite{xu2022opv2v} training pipeline to develop the cooperative perception baselines.

\parab{Backbone architecture.}
We adopt the PointPillars backbone configured in OpenCOOD. Each LiDAR sweep is voxelized with a resolution of 0.4 m in both x and y directions, with up to 32 points per voxel and a maximum of 32k non-empty voxels per frame. The pillar encoder extracts per-voxel features, which are then processed by a 2D CNN to produce bird’s-eye-view (BEV) feature maps for subsequent stages.

\parab{Fusion strategies.}
We implement four cooperative perception baselines, including V2VNet, V2X-ViT, V2VAM and CoBEVT. Intermediate-feature-level models exchange BEV features after spatial alignment using the ground-truth poses provided in COOPERSCENE.

\parab{Training schedule.}
We adopt a batch size of 4 on a single GPU server equipped with NVIDIA RTX A6000 GPUs. Each model is trained for around 40 epochs. Data augmentation follows OpenCOOD defaults, including random flipping along the x-axis, random rotation within ±45°, and global scaling.

\section{Labeling Tools}

\begin{figure*}[t]
    \centering
    \includegraphics[width=\textwidth]{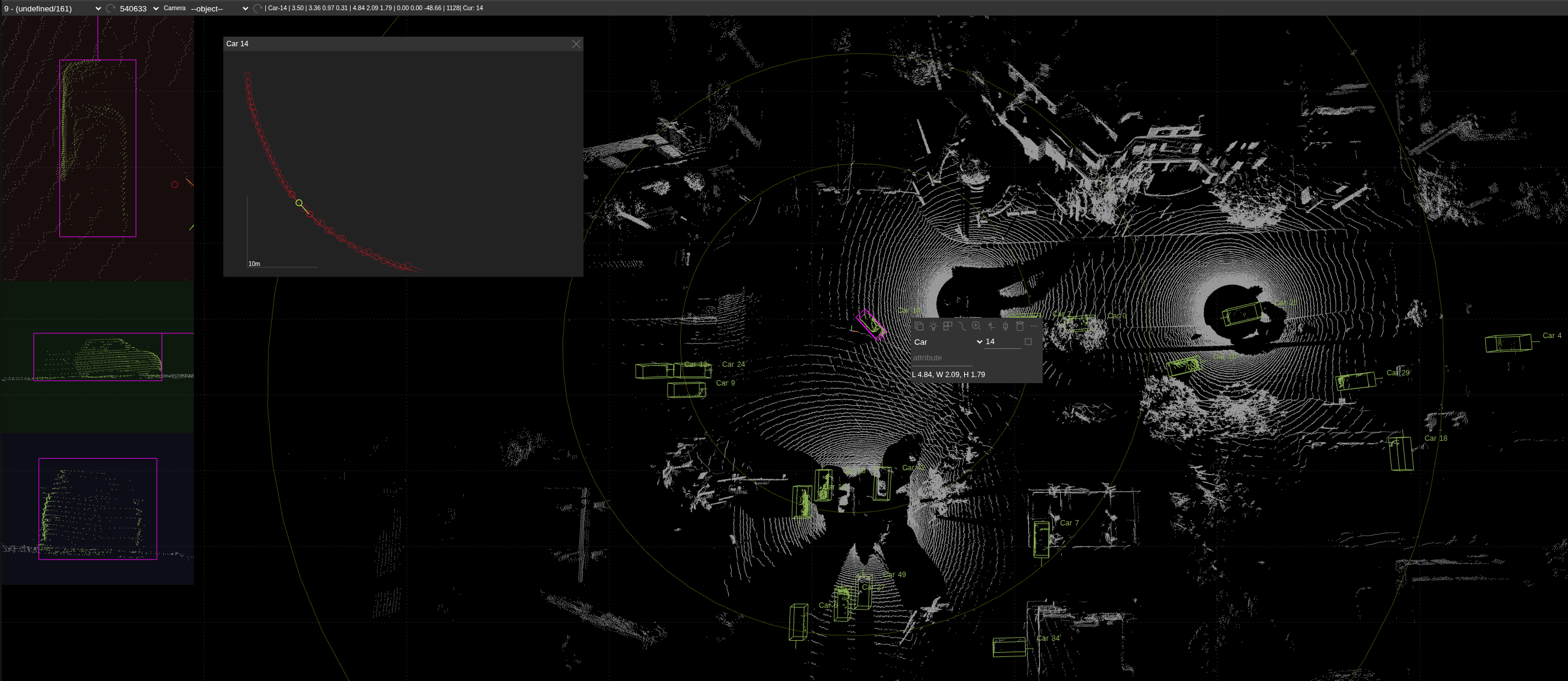}
    \caption{SUSTechPoints annotation interface for human refinement. The fused LiDAR point cloud from three CAVs and road side setup is loaded with the global bounding boxes obtained from the auto labeling stage. The SUSTechPoints tool is able to visualize LiDAR sweeps, 3D bounding boxes, and review object trajectories across time. The interface provides multi-view panels for fine-grained adjustment of object orientation and size, supporting high-quality correction of bounding boxes.}
    \label{fig:sustechpoints_1}
\end{figure*}

\begin{figure*}[t]
    \centering
    \includegraphics[width=\textwidth]{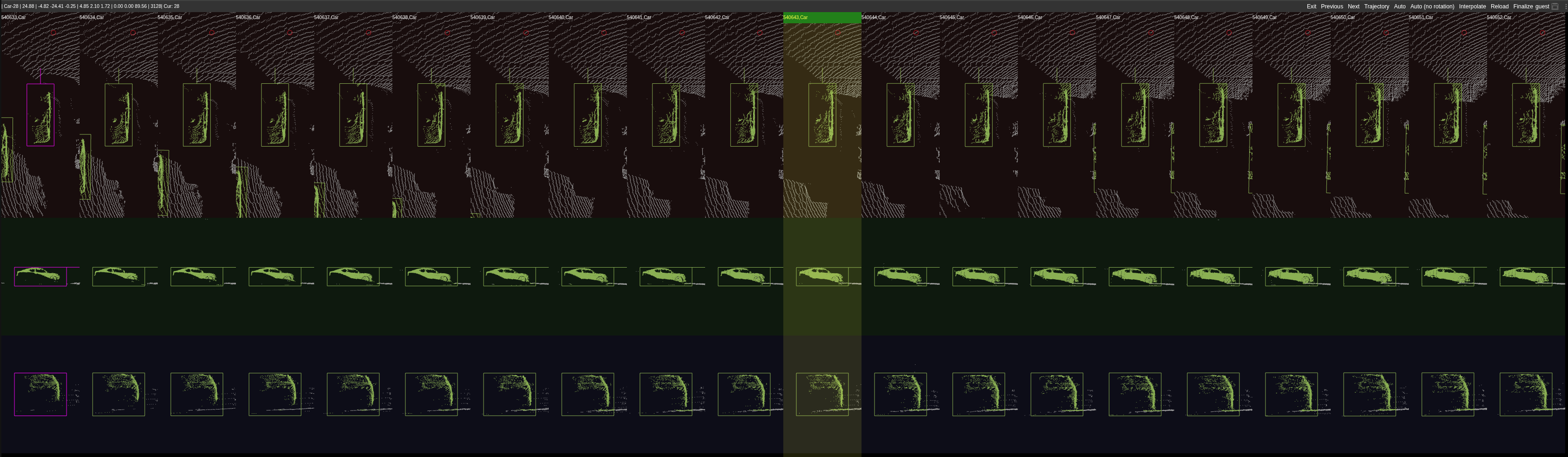}
    \caption{Cross-frame trajectory adjustment in SUSTechPOINTS. SUSTechPOINTS provides a multi-frame visualization interface that allows annotators to refine a single object’s 3D bounding box consistently across time. Annotators can adjust the bounding box in any frame and propagate corrections using built-in interpolation tools, which automatically generate smooth and temporally coherent trajectories.}
    \label{fig:sustechpoints_2}
\end{figure*}

To efficiently annotate large-scale multi-agent LiDAR data, \sysname adopts a hybrid auto-labeling + human refinement workflow. As introduced in Section 3, we first apply BEVFusion detector to generate initial 3D bounding box proposals for all LiDAR sweeps. The detected boxes are then linked across frames using the AB3DMOT tracker to produce temporally consistent object trajectories. Then we apply late fusion module to fuse the bounding boxes across agents, and finally the processed labels are refined using the SUSTechPOINTS labeling tool ~\cite{sustecth_points}.

Figures~\ref{fig:sustechpoints_1} and~\ref{fig:sustechpoints_2} showcase the SUSTechPOINTS labeling interface, where annotators can inspect generated 3D bounding boxes, correct geometric inaccuracies, and ensure cross-frame spatial alignment. SUSTechPOINTS provides a rich set of functionalities including interactive editing for refining 3D bounding boxes, and trajectory-aware labeling. These capabilities are particularly important for \sysname, where multi-agent interactions and occlusions require consistent annotations that span several perspectives.

\section{Representative Driving Scenarios}

\definecolor{strongpurple}{RGB}{128, 0, 128}

\begin{figure}[!b]
    \centering
    \caption{LiDAR overlay visualization across multiple traffic scenarios involving three connected automated vehicles (CAVs) and one roadside infrastructure unit. Infrastructure LiDAR is shown in \textcolor{strongpurple}{purple} points, while CAV A, CAV B, and CAV C are visualized in \textcolor{red}{red}, \textcolor{green}{green}, and \textcolor{blue}{blue}, respectively.}
    \label{fig:driving_scenarios}
    \setlength{\tabcolsep}{3pt}

    \begin{tabular}{cc}
        \begin{subfigure}{0.45\textwidth}
            \centering
            \includegraphics[width=\linewidth]{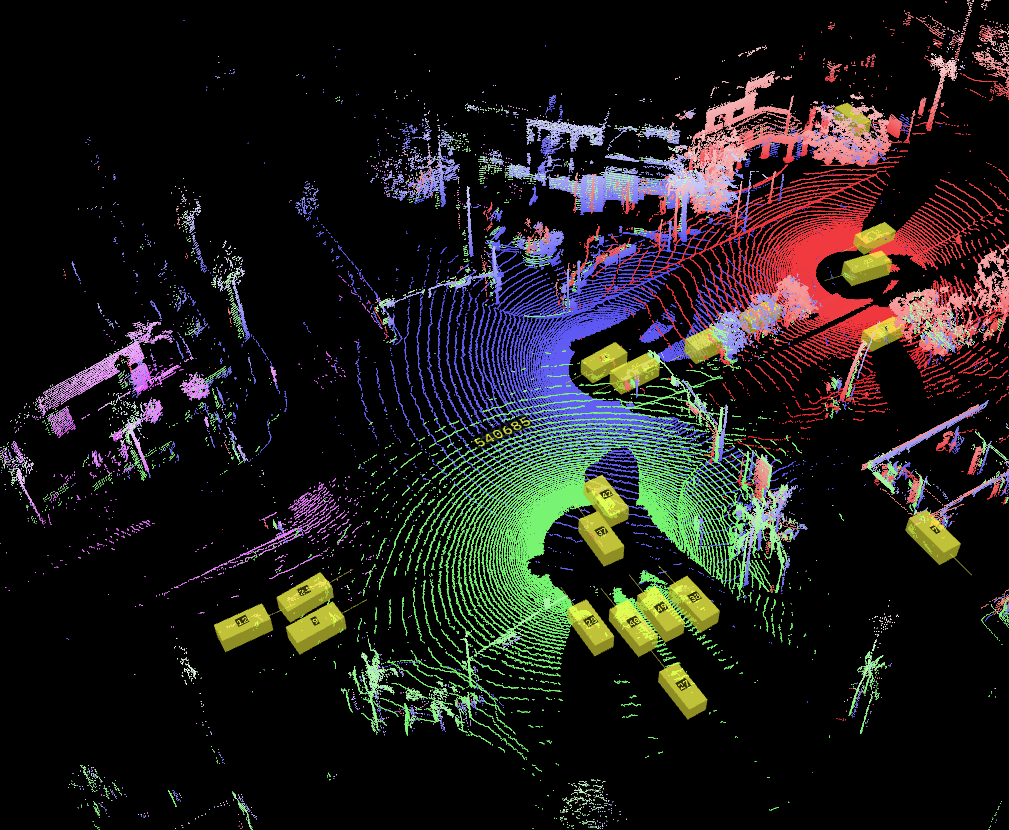}
            \caption{Scenario 1}
        \end{subfigure} &
        \begin{subfigure}{0.45\textwidth}
            \centering
            \includegraphics[width=\linewidth]{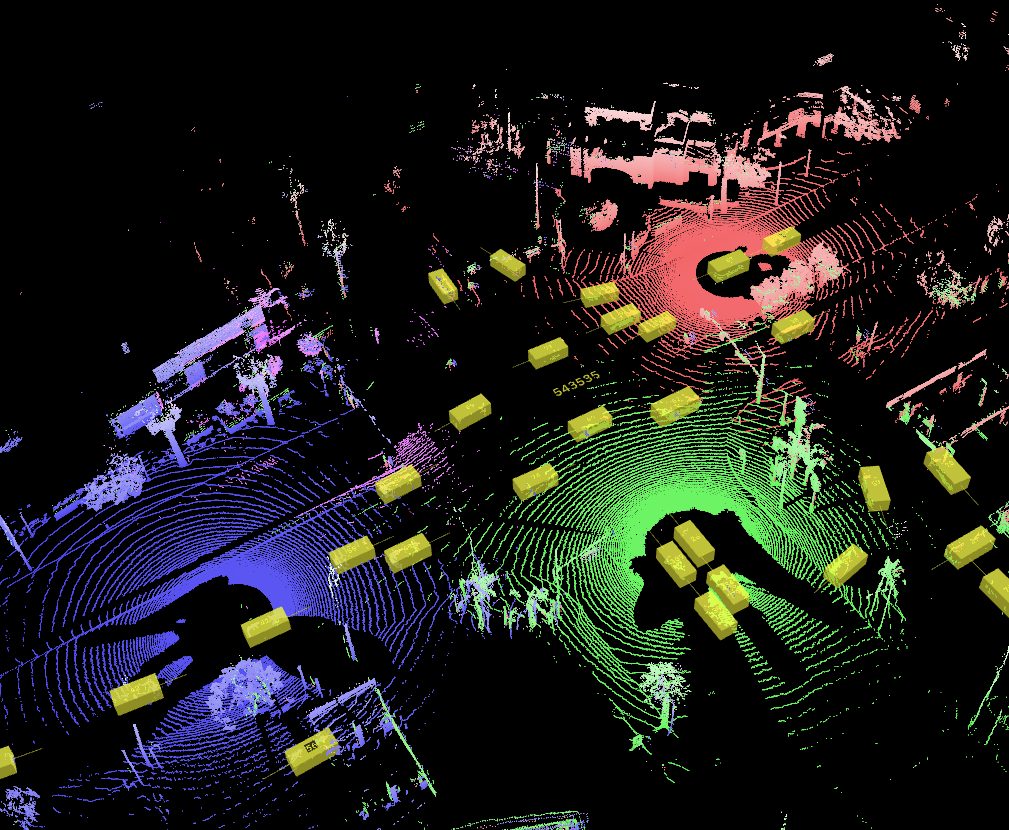}
            \caption{Scenario 2}
        \end{subfigure} \\

        \begin{subfigure}{0.45\textwidth}
            \centering
            \includegraphics[width=\linewidth]{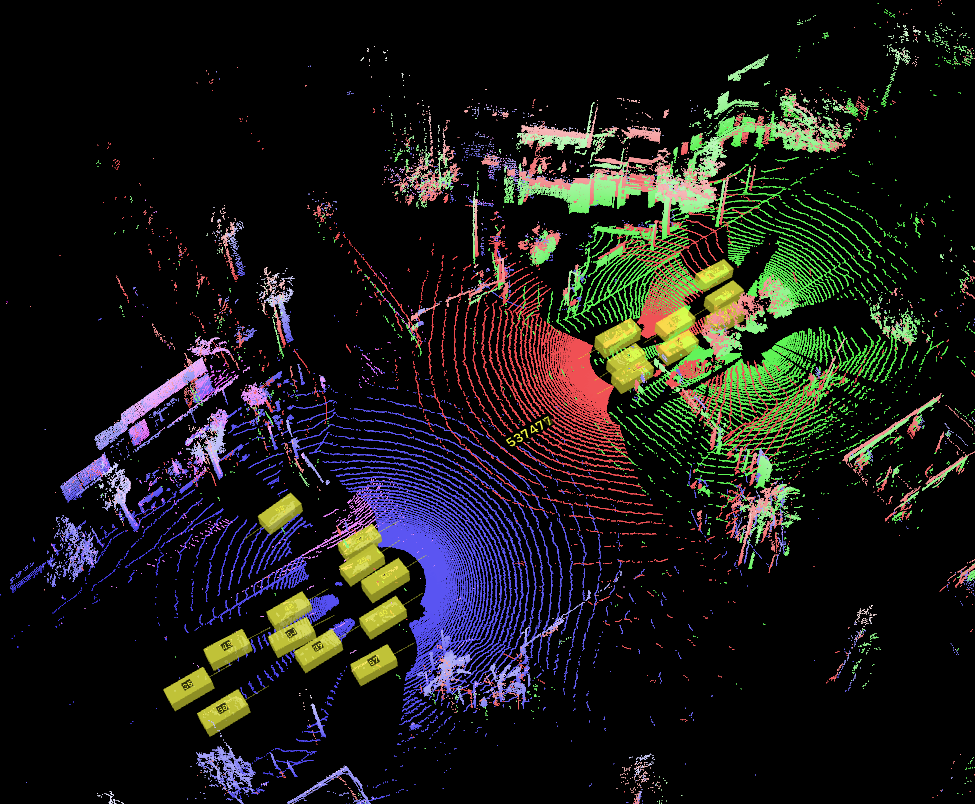}
            \caption{Scenario 3}
        \end{subfigure} &
        \begin{subfigure}{0.45\textwidth}
            \centering
            \includegraphics[width=\linewidth]{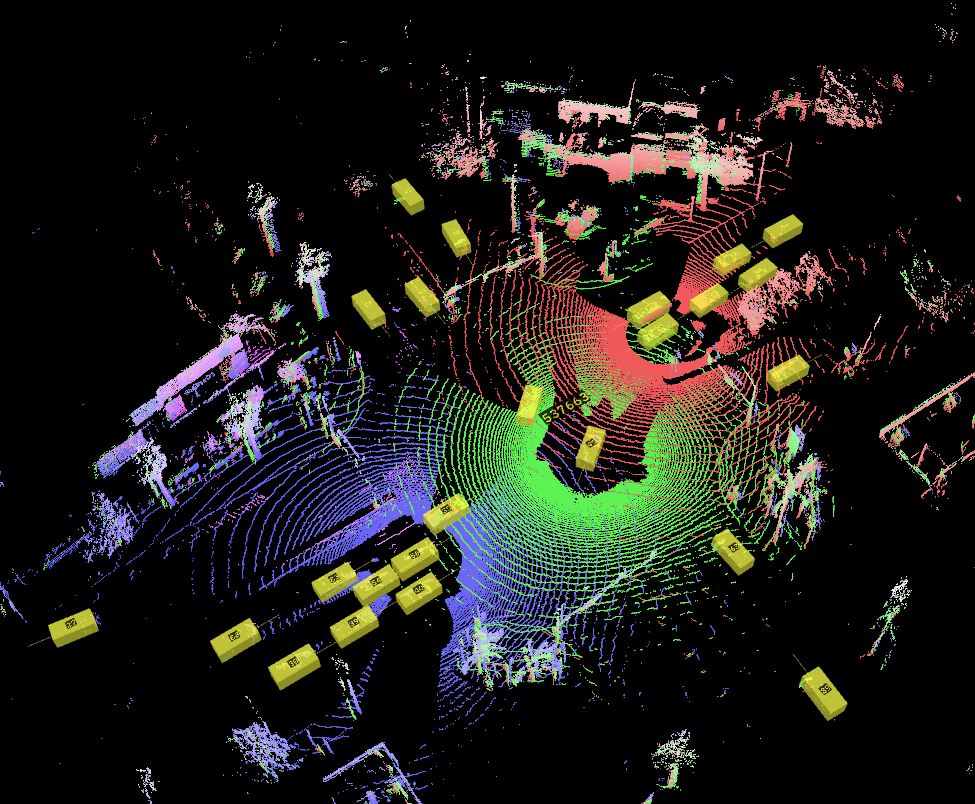}
            \caption{Scenario 4}
        \end{subfigure} \\

        \begin{subfigure}{0.45\textwidth}
            \centering
            \includegraphics[width=\linewidth]{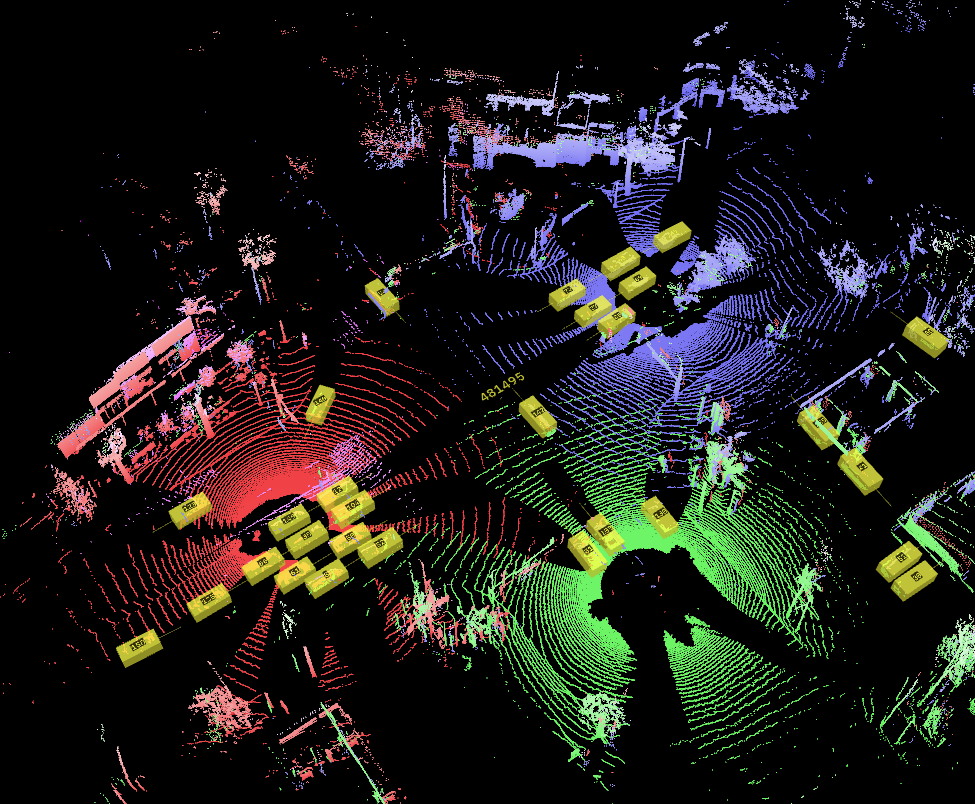}
            \caption{Scenario 5}
        \end{subfigure} &
        \begin{subfigure}{0.45\textwidth}
            \centering
            \includegraphics[width=\linewidth]{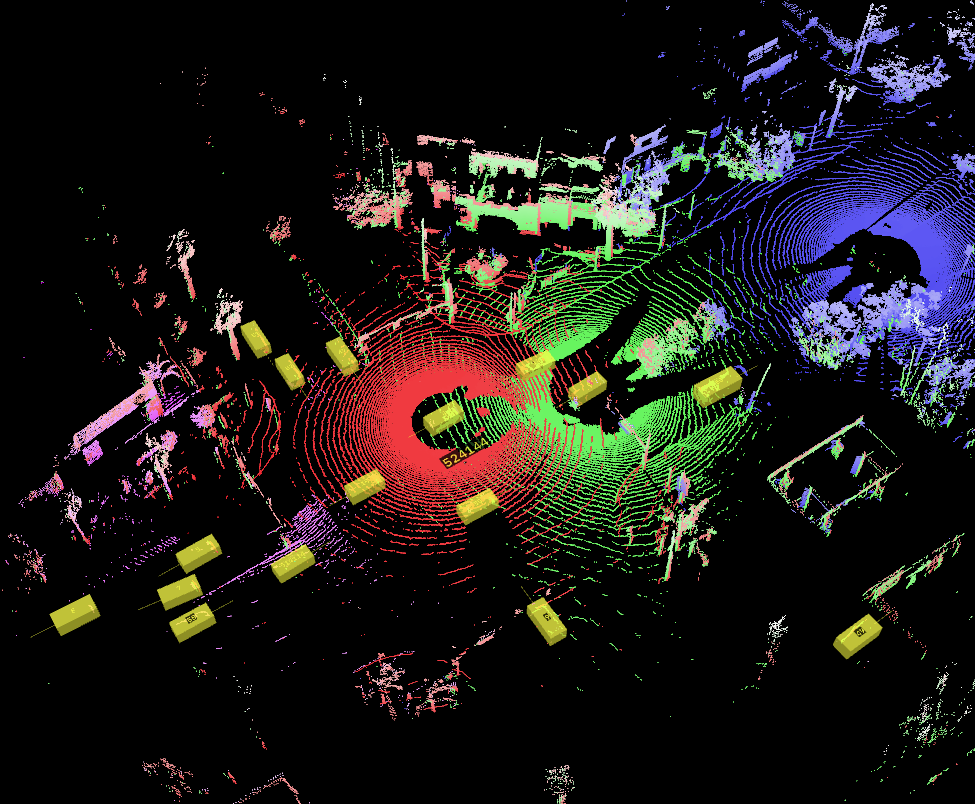}
            \caption{Scenario 6}
        \end{subfigure} \\
    \end{tabular}
\end{figure}

\sysname covers a wide spectrum of real-world cooperative driving situations involving 
multi-agent interactions around a busy urban intersection. The dataset captures diverse traffic dynamics, including vehicles entering and exiting the intersection, partial and full occlusions, long-distance separation between agents, complementary viewpoints. These rich scenarios illustrate the practical challenges and opportunities for cooperative perception and serve as an important testbed for evaluating cross-agent sensing and reasoning.

To provide a clearer understanding of these multi-agent interactions, this section highlights six representative driving scenarios from \sysname. Each example demonstrates how different CAVs and the roadside infrastructure contribute complementary observations under various conditions. Figure~\ref{fig:driving_scenarios} shows LiDAR overlay visualizations for these scenarios.

\vspace{-16pt}
\subsection*{Scenario 1}\vspace{-8pt}
CAV B(green) is going through the intersection, but its right-hand-side view is occluded by nearby moving vehicles. This missing perspective is effectively compensated by the observations from the other two CAVs.

\vspace{-16pt}
\subsection*{Scenario 2}\vspace{-8pt}
CAV C (blue) is approaching the intersection but remains too far from CAV A (red) for observation of each other's surrounding environment. However, both vehicles benefit from cooperative information relayed by CAV B (green), which provides complementary viewpoints to compensate for their limited overlap.

\vspace{-16pt}
\subsection*{Scenario 3}\vspace{-8pt}
CAV B (green) has its front view partially occluded by vehicles ahead, preventing it from observing the opposite side of the intersection. This missing visibility is compensated by CAV A (red).

\vspace{-16pt}
\subsection*{Scenario 4}\vspace{-8pt}
CAV B (green) is positioned near the center of the intersection, providing broad 360-degree visibility and clear observations of surrounding traffic from all directions.

\vspace{-16pt}
\subsection*{Scenario 5}\vspace{-8pt}
The three CAVs approach the intersection from different directions, each contributing their perspective view. Their complementary points provide comprehensive coverage of the scenario.

\vspace{-16pt}
\subsection*{Scenario 6}\vspace{-4pt}
CAV C (blue) is exiting the intersection, moving out of the shared coverage region. CAV A (red) loses cooperative overlap with CAV C, reducing their joint visibility.


\end{document}